%% file: main.tex
\documentclass[10pt]{article} 
\usepackage[accepted]{tmlr}

\input{math_commands.tex}

\usepackage[export]{adjustbox}
\usepackage{amsfonts}       
\usepackage{amsmath}
\usepackage{amssymb}
\usepackage[toc,page]{appendix}
\usepackage{array}
\usepackage{booktabs}       
\usepackage{caption}
\usepackage{enumitem}
\usepackage[T1]{fontenc}    
\usepackage{graphicx}
\usepackage{hyperref}
\usepackage[utf8]{inputenc} 
\usepackage{mathtools}
\usepackage{microtype}      
\usepackage{multirow}
\usepackage{nicefrac}       
\usepackage{url}            
\usepackage{wrapfig}
\usepackage[table, dvipsnames]{xcolor}         
\usepackage{xspace}

\newcommand{\ourmethod}{LightKV}

\newcommand{\eg}{\emph{e.g.}\@}

\makeatletter
\DeclareRobustCommand\onedot{\futurelet\@let@token\@onedot}
\def\@onedot{\ifx\@let@token.\else.\null\fi\xspace}
\makeatother

\title{Make Your LVLM KV Cache More Lightweight}

\author{\name Xihao Chen 
        \email chenxihao@u.nus.edu \\
        \addr Integrative Sciences and Engineering Programme, National University of Singapore \\
        \addr School of Computing, National University of Singapore
        \AND
        \name Yangyang Guo\thanks{Corresponding author.}
        \email guoyang.eric@gmail.com \\
        \addr School of Computing, National University of Singapore
        \AND
        \name Roger Zimmermann \email dcsrz@nus.edu.sg \\
        \addr School of Computing, National University of Singapore
}

\def\month{04}  
\def\year{2026} 
\def\openreview{\url{https://openreview.net/forum?id=n77IeySrQl}} 

\begin{document}

\maketitle

\input{sec/0_abstract}
\input{sec/1_intro}
\input{sec/2_related_work}
\input{sec/3_method}
\input{sec/4_experiments}
\input{sec/5_conclusion}

\bibliography{main}
\bibliographystyle{tmlr}

\clearpage
\appendix
\input{sec/X_suppl}

\end{document}

%% file: math_commands.tex
\usepackage{amsmath,amsfonts,bm}

\def\eqref#1{equation~\ref{#1}}

\def\floor#1{\lfloor #1 \rfloor}
\def\1{\bm{1}}

\DeclareMathAlphabet{\mathsfit}{\encodingdefault}{\sfdefault}{m}{sl}
\SetMathAlphabet{\mathsfit}{bold}{\encodingdefault}{\sfdefault}{bx}{n}



%% file: sec/0_abstract.tex
\begin{abstract}
Key-Value (\textbf{KV}) cache has become a \emph{de facto} component of modern Large Vision-Language Models (\textbf{LVLM}s) for inference.
While it enhances decoding efficiency in Large Language Models (\textbf{LLMs}), its direct adoption in LVLMs introduces substantial GPU memory overhead due to the large number of vision tokens processed during the prefill stage.
To tackle this problem, we propose \ourmethod, a novel approach that reduces KV cache size by exploiting the redundancy among vision-token embeddings.
Guided by text prompts, \ourmethod\ employs cross-modality message passing to aggregate informative messages across vision tokens and progressively compress them during prefill.
This prompt-aware guidance distinguishes our method from prior vision-only compression strategies.
We evaluate \ourmethod\ on eight open-source LVLMs across eight public benchmark datasets, \eg, MME and SeedBench.
Experimental results demonstrate that with only 55\% of the original vision tokens, \ourmethod\ (a) halves the vision-token KV cache size, (b) reduces computation by up to 40\%, and (c) preserves general-purpose performance while significantly outperforming existing baselines.
Our code is publicly available at \href{https://github.com/howtoosee/LightKV}{https://github.com/howtoosee/LightKV}.
\end{abstract}

%% file: sec/1_intro.tex
\section{Introduction}
\label{sec:intro}

Benefiting from the rapid advancements in Large Language Models (LLMs)~\citep{thevicunateamVicunaOpenSourceChatbot2023, openaiGPT4TechnicalReport2024, llamateamLlama3Herd2024}, Large Vision-Language Models (LVLMs)~\citep{alayracFlamingoVisualLanguage2022, liBLIP2BootstrappingLanguageImage2023, daiInstructBLIPGeneralpurposeVisionLanguage2023, baiQwenVLVersatileVisionLanguage2023, liuVisualInstructionTuning2023, liuImprovedBaselinesVisual2024, liuLLaVANeXTImprovedReasoning2024, luDeepSeekVLRealWorldVisionLanguage2024, chenInternVLScaling2024, chenHowFarAre2024, wangEnhancingReasoningAbility2025, chenExpandingPerformanceBoundaries2025} have recently garnered extensive attention.
For example, LLaVA~\citep{liuVisualInstructionTuning2023} and DeepSeek-VL~\citep{luDeepSeekVLRealWorldVisionLanguage2024} have achieved impressive performance on a multitude of general-purpose multi-modal benchmarks~\citep{fuMMEComprehensiveEvaluation2024, yuMMVetEvaluatingLarge2024, liEvaluatingObjectHallucination2023}.
However, the efficiency of LVLMs remains a significant bottleneck for researchers and practitioners in resource-constrained environments.

Key-Value (\textbf{KV}) cache~\citep{popeEfficientlyScalingTransformer2023, kwonEfficientMemoryManagement2023} serves as a fundamental technique in optimizing the inference efficiency of mainstream LLMs and LVLMs. 
However, although KV caching improves inference speed without compromising model performance, it substantially increases GPU memory consumption.
This limitation is especially severe with longer sequences generated~\citep{yangPyramidInferPyramidKV2024, liuMiniCacheKVCache2024, liSnapKVLLMKnows2024}.
To alleviate this issue, some training-based methods, such as MQA~\citep{huMatryoshkaQueryTransformer2025} and GQA~\citep{ainslieGQATrainingGeneralized2023}, introduce the sharing of keys and values across different attention heads.
As such, the overall KV cache size is accordingly reduced.
These approaches, however, suffer from the requirement of heavy model retraining.
In contrast, other methods, such as H2O~\citep{zhangH2OHeavyHitterOracle2023}, MiniCache~\citep{liuMiniCacheKVCache2024}, and ElasticCache~\citep{liuEfficientInferenceVision2024} focus on pruning tokens within the KV cache \textit{during inference} after the prefill stage. 
These methods offer greater flexibility and can be seamlessly applied to existing decoder-only LVLM models with minimal degradation in performance.
\textit{Given this, our work primarily focuses on the reduction of vision tokens during inference time.}

\begin{figure}[th]
    \centering
    \includegraphics[width=0.6\linewidth]{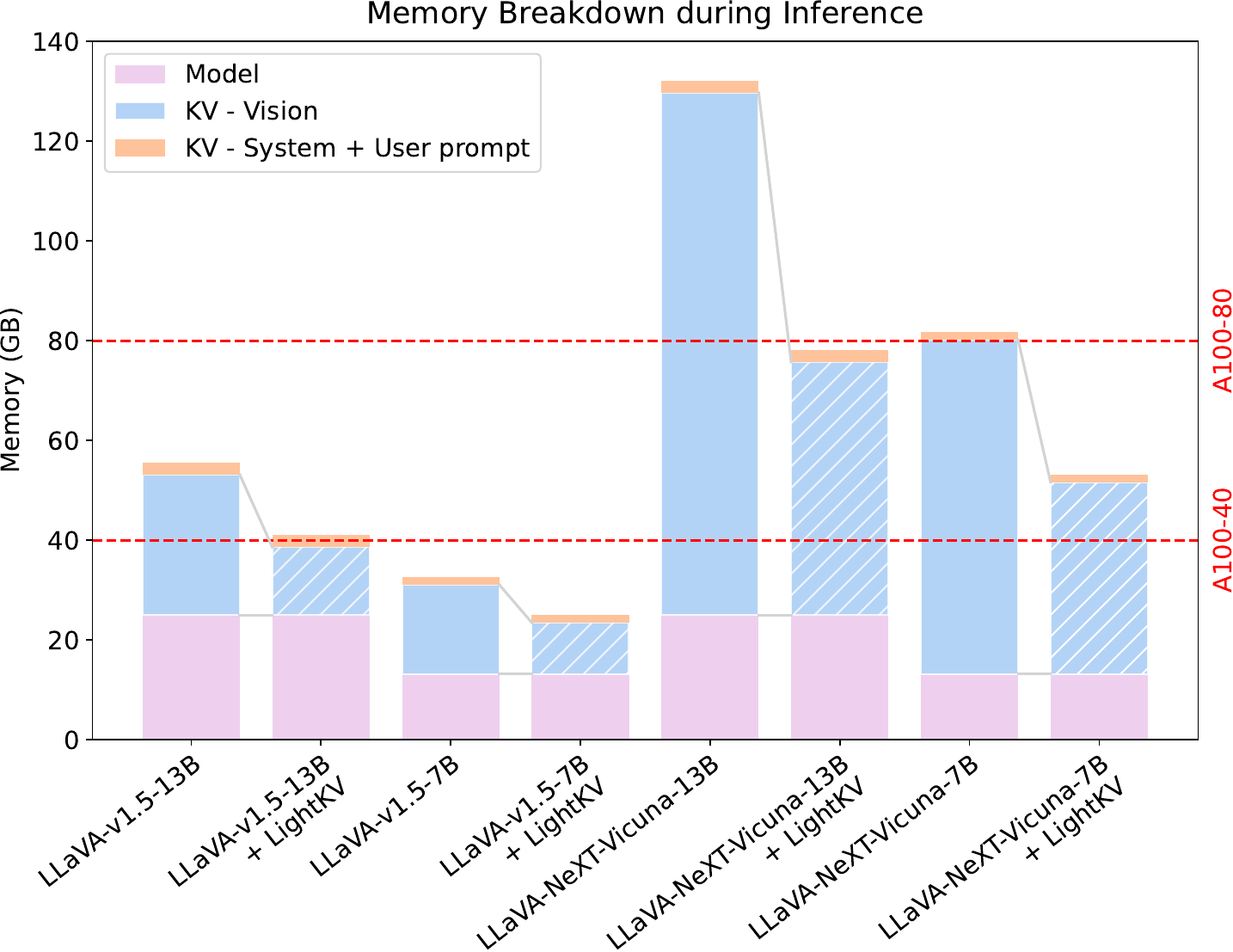}
    \caption{Breakdown of memory consumption in LLaVA models during prefill shows the substantial reduction in KV cache usage with \ourmethod. As LLaVA-NeXT uses approximately $4\times$ the \textit{vision tokens} as LLaVA-v1.5, there is a sharp increase in memory consumption.}
    \label{fig:memory-breakdown}
    \vspace{-1em}
\end{figure}

Unlike LLMs, reducing the cost of memory-bound KV cache is challenging in LVLMs due to the following two factors:
(a) Tokens in LVLMs are heterogeneous, representing both image patches and text.
Determining which tokens should be pruned thus becomes more difficult;
(b) The number of tokens computed during the \textit{prefill stage} is significantly larger than that in LLMs.
Each image or video frame in LVLMs is embedded into hundreds to thousands of tokens upfront (\eg, 576 in LLaVA-v1.5~\citep{liuVisualInstructionTuning2023} and 7,290 in LLaVA-OneVision~\citep{liLLaVAOneVisionEasyVisual2024}), a considerable amount compared to the context lengths of LLMs (see Fig.~\ref{fig:memory-breakdown})~\citep{llamateamLlama3Herd2024, jiangMistral7B2023, thevicunateamVicunaOpenSourceChatbot2023}.
As a result, current LVLMs are limited by significantly heavier GPU memory usage than their LLM counterparts during prefill.
A few recent studies have proposed addressing the first challenge on token heterogeneity~\citep {chenEfficientLargeMultimodal2024, liLLaMAVIDImageWorth2024}.
However, existing research on solving the second challenge remains sparse.

In this paper, we propose \ourmethod, a novel method for optimizing KV cache storage in LVLMs during the prefill stage \textbf{without retraining}.
To this end, we leverage cross-modal prompt guidance to compress vision tokens.
Our method follows a three-step design.
\textit{First}, we conceptually map each vision token to a graph node, constructing a bipartite graph with edges representing a feature divergence (FD) metric between the connected nodes.
Nonetheless, computing FD in a pairwise manner is still expensive, especially with a large number of vision tokens.
To alleviate this problem, \textit{second}, we split the vision tokens into sub-windows based on their original spatial locations. 
This allows us to reduce the complexity of computing FD and aggregating information across tokens, thus improving efficiency.
\textit{Third}, our method does not follow existing studies~\citep{chenImageWorth12024} to perform vision token reduction independently, as the text prompts offer more informative signals for vision token importance.
Consequently, we leverage on-the-fly cross-modal attention scores between vision tokens and prompt tokens for informed token updates.
We find that although this approach has been largely ignored by the existing literature, it delivers superior results to state-of-the-art baselines. 

We apply \ourmethod\ to eight state-of-the-art LVLM models: 
LLaVA-v1.5-13B,
LLaVA-v1.5-7B~\citep{liuVisualInstructionTuning2023},
LLaVA-NeXT-13B,
LLaVA-NeXT-7B~\citep{liuImprovedBaselinesVisual2024},
InternVL2-8B~\citep{chenHowFarAre2024},
EVE-7B-v1,
EVE-7B-v1-HD~\citep{diaoUnveilingEncoderfreeVisionlanguage2025},
Qwen2.5-VL~\citep{baiQwen25VLTechnicalReport2025},
and conduct extensive experiments across eight benchmarks, \eg,
MME~\citep{fuMMEComprehensiveEvaluation2024} and 
SeedBench~\citep{liSEEDBenchBenchmarkingMultimodal2024}.
Our results demonstrate that \ourmethod\ can reduce the KV memory of vision tokens by $50\%$ while maintaining, sometimes even surpassing, the vanilla LVLM performance.
Furthermore, when constrained with the same token length generation budget, the inference cost in FLOPs is reduced by 40\%.

In summary, \ourmethod\ reduces the KV cache footprint in LVLMs by compressing vision tokens during the \textit{prefill} stage under the guidance of text prompts.
This prompt-aware design distinguishes it from existing SOTA vision-only methods, delivering (1) greater efficiency and (2) superior benchmark performance.
Importantly, \ourmethod\ is entirely \textit{training-free} and can be seamlessly applied to a wide range of LVLMs, including both vision encoder-based and encoder-free models.

%% file: sec/2_related_work.tex
\section{Related work}
\label{sec:relatedwork}

\paragraph{Large vision-language models}
Following the success of large language models (LLMs) in the language domain~\citep{thevicunateamVicunaOpenSourceChatbot2023, openaiGPT4TechnicalReport2024, llamateamLlama3Herd2024}, large vision-language models (LVLMs) have shown substantial progress on various multimodal tasks~\citep{geminiteamGeminiFamilyHighly2024, geminiteamGemini15Unlocking2024, driessPaLMEEmbodiedMultimodal2023}.
Current LVLMs primarily fall into the following three directions:
(a) Fusion-based methods directly inject vision information into the LLM decoders using cross-attention~\citep{alayracFlamingoVisualLanguage2022, awadallaOpenFlamingoOpenSourceFramework2023, liOtterHDHighResolutionMultimodality2023, gongMultiModalGPTVisionLanguage2023}.
(b) Query-based LVLMs extract vision information with learnable query tokens, which are then concatenated with text tokens~\citep{liBLIP2BootstrappingLanguageImage2023, daiInstructBLIPGeneralpurposeVisionLanguage2023, zhuMiniGPT4EnhancingVisionLanguage2024, liLLaMAVIDImageWorth2024, zhangVideoLLaMAInstructiontunedAudioVisual2023}.
(c) Projection-based methods directly map the encoded tokens from a vision encoder into the text space~\citep{liuVisualInstructionTuning2023, liuImprovedBaselinesVisual2024, liuLLaVANeXTImprovedReasoning2024, liLLaVAOneVisionEasyVisual2024, baiQwenVLVersatileVisionLanguage2023, huangLanguageNotAll2023, diaoUnveilingEncoderfreeVisionlanguage2025}.
However, despite their simplicity, such a projection substantially increases the memory footprint of the input sequence.

\paragraph{KV cache optimization}
KV cache has been widely used in LLMs and LVLMs to improve their inference efficiency~\citep{daoFlashAttentionFastMemoryEfficient2022, popeEfficientlyScalingTransformer2023, kwonEfficientMemoryManagement2023, leeInfiniGenEfficientGenerative2024}.
The core idea is to store the key and value tokens to reduce future redundant computations.
However, in situations with long contexts, keeping the KV cache imposes an increased burden on GPU memory.
Existing approaches addressing this can be roughly categorized into two groups: (a) KV-sharing-based and (b) token-reduction-based.
Specifically, methods from (a) improve the multi-headed attention mechanism to achieve efficiency.
For instance, MQA~\citep{huMatryoshkaQueryTransformer2025} and GQA~\citep{ainslieGQATrainingGeneralized2023} share keys and values across attention heads~\citep{vaswaniAttentionAllYou2017}, reducing the amount of KV needed to be cached.
In contrast, methods from (b) improve KV cache size by pruning or merging tokens based either on minimal importance~\citep{zhangH2OHeavyHitterOracle2023, liSnapKVLLMKnows2024, caiPyramidKVDynamicKV2024} or attention consistency across layers~\citep{liuScissorhandsExploitingPersistence2023, liuEfficientInferenceVision2024, yangPyramidInferPyramidKV2024}.
Beyond LLMs, some initial efforts have been devoted to optimizing the KV cache for LVLMs.
In particular, LLaVolta~\citep{chenEfficientLargeMultimodal2024}, IVTP~\citep{huangIVTPInstructionguidedVisual2024} and FastV~\citep{chenImageWorth12024} propose pruning vision tokens in the LLM decoder backbone.
The first two require model retraining; FastV, though training-free, prunes vision tokens without cross-modality guidance, yielding inconsistent results across models and benchmarks.
In contrast, \ourmethod\ leverages guidance from text tokens to deliver more consistent and superior performance across a diverse set of benchmarks.

\paragraph{Vision token compression}
Tokens in vision transformers (ViTs)~\citep{dosovitskiyImageWorth16x162021} often exhibit high redundancy~\citep{bolyaTokenMergingYour2023, panLessMorePay2022, chenImageWorth12024}.
To address this, some approaches train modules to identify and discard less important tokens~\citep{raoDynamicViTEfficientVision2021, bonnaerensLearnedThresholdsToken2023, yinAViTAdaptiveTokens2022, fayyazAdaptiveTokenSampling2022, weiJointTokenPruning2023, chenDiffRateDifferentiableCompression2023, zhangLLaVAMiniEfficientImage2024, maoPruneMergeEfficient2025}.
Some other typical methods first group tokens based on similarity or distance~\citep{bolyaTokenMergingYour2023, tranAcceleratingTransformersSpectrumPreserving2024, kimTokenFusionBridging2024, alvarDivPruneDiversitybasedVisual2025} or image segmentation~\citep{xuGroupViTSemanticSegmentation2022, luContentawareTokenSharing2023} and then prune or merge the tokens with the maximum similarity.
These methods either (a) require the training of additional module(s), or (b) do not support the vision-language joint reasoning as in LVLMs.

%% file: sec/3_method.tex
\begin{figure*}[t!]
    \centering
    \includegraphics[width=\linewidth]{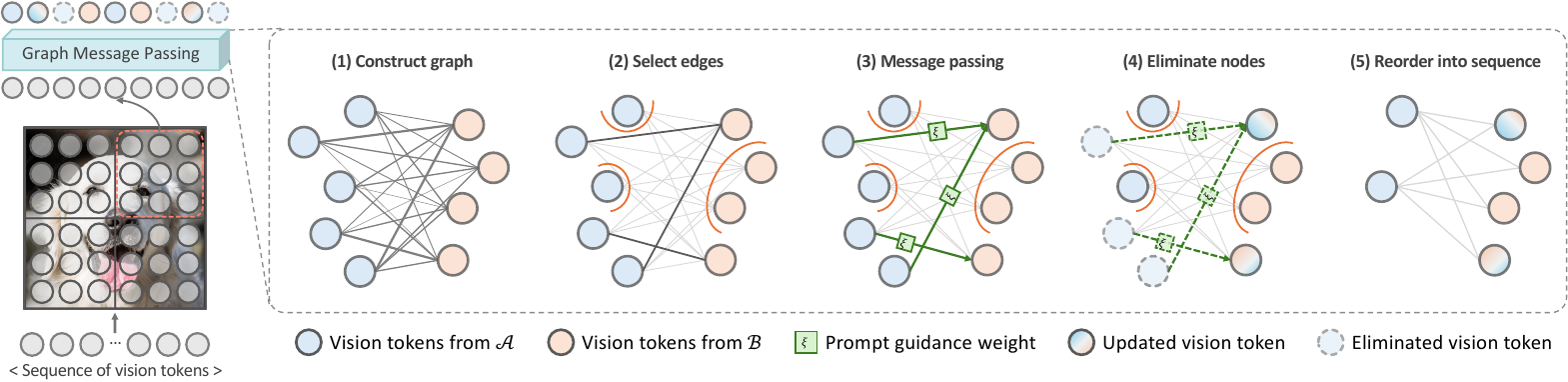}
    \caption{Method overview of intra-window token compression.
    \textbf{Step 1}: Construct a bipartite graph by partitioning the vision tokens into non-overlapping sets $\mathcal{A}$ (\textcolor{blue!60}{blue}) and $\mathcal{B}$ (\textcolor{orange!70}{orange}), weight each edge by an FD metric, defined in Eq.~\ref{eq:similarity}. 
    \textbf{Step 2}: Select edges with the smallest $\floor{\rho v/2}$ FD values and delete the rest. The unconnected nodes are left unchanged.
    \textbf{Step 3}: Pass messages from nodes in $\mathcal{A}$ to connected nodes in $\mathcal{B}$, weighted by their corresponding attention scores $\xi$, as computed in Eq.~\ref{eq:attention-factor}. Then, aggregate messages and update nodes in $\mathcal{B}$.
    \textbf{Step 4}: Eliminate the now-redundant nodes from $\mathcal{A}$.
    \textbf{Step 5}: Reorder the remaining nodes into a sequence of vision tokens, serving as input to the next decoder layer.}
    \label{fig:method-overview}
    \vspace{-1em}
\end{figure*}

\section{Method}
\label{sec:method}

\subsection{Preliminaries}
\label{sec:preliminaries}

Recent LLMs often operate in an autoregressive fashion: given a sequence of $p$ text prompt tokens $\left[x_1, \ldots, x_{p} \right]$ (including both system prompt and user prompt), and $t-p$ previously generated tokens $\left[x_{p+1}, \ldots, x_{t} \right]$, an LLM with parameters $\Theta$ predicts the next token $x_{t+1}$ with:
\begin{equation}
    \label{eq:llm-generation}
    \mathbb{P}_{\Theta} \big(x_{t+1} \ | \ \underbrace{x_1, \ldots, x_{p}}_{\text{Prompt tokens}}, \underbrace{x_{p+1}, \ldots, x_t}_{\text{Generated tokens}} \big).
\end{equation}

The above process is often implemented in two stages: prefill and generation~\citep{goldenGenerativeAILLMs2024}.
During \emph{prefill}, the model tokenizes all $p$ prompt tokens and computes the queries $Q_{p} = \left[ \mathbf{q}_1, \mathbf{q}_2, \ldots, \mathbf{q}_{p} \right]$, similarly for keys $K_{p}$ and values $V_{p}$~\citep{vaswaniAttentionAllYou2017}.
In contrast, during \emph{generation}, when a new token arrives, the model first obtains the query $\mathbf{q}_{t+1}$, key $\mathbf{k}_{t+1}$, and value $\mathbf{v}_{t+1}$ vectors.
It then computes the attention matrix by applying $\mathbf{q}_{t+1}$ to the full set of keys $K_{t+1}$:
\begin{equation}
    \label{eq:attention-matrix}
    \mathbf{A} = \operatorname{softmax} \left( {\mathbf{q}_{t+1}\ K^{\top}_{t+1}} / {\sqrt{d_k}} \right),
\end{equation}
where $d_k$ represents the embedding dimension.
In practice, the attention output would be a concatenation of matrices $\mathbf{A} = [\mathbf{A}_1, \ldots, \mathbf{A}_H]$ from $H$ independent attention heads.

\paragraph{KV cache}
From the above, we observe that the autoregressive nature of LLMs allows for the previously computed \textit{keys} $K_t$ and \textit{values} $V_t$ to be reused in future time steps during generation.
This operation reduces the computational overhead by preventing the recomputation of key and value tokens~\citep{xuFastOndeviceLLM2025}.
However, an increased consumption of GPU memory is usually induced by the growing size of the KV cache.
This is often manifested as: (a) generating lengthy sequences and (b) caching many contexts during prefill.
In this work, we primarily focus on improving the second.

\paragraph{LVLMs} LVLMs build on LLMs by extending their architecture to process visual information.
A common paradigm in LVLMs is to first map the split image patches into tokens using ViT-based encoders~\citep{dosovitskiyImageWorth16x162021, radfordLearningTransferableVisual2021, baoBEiTBERTPreTraining2022}, and then concatenate these tokens with the prompt tokens to form the input sequence.
In general, LVLMs generate tokens by conditioning on both text prompt tokens and vision tokens:
\begin{equation}
    \label{eq:lvlm-generation}
    \mathbb{P}_{\Theta} \big(x_{t+1} \ | \ \underbrace{x_1, \ldots, x_{p}}_{\text{Prompt tokens}}, \underbrace{x_{p+1}, \ldots, x_{p+v}}_{\text{Vision tokens}}, \underbrace{x_{p+v+1}, \ldots, x_t}_{\text{Generated tokens}} \big).
\end{equation}
We denote $X_\mathtt{v}$ as the sequence of $v$ vision tokens in Eq.~\ref{eq:lvlm-generation}.
Similar to LLMs, KV cache is a key component in speeding up inference in LVLMs.
In this paper, we focus primarily on compressing vision tokens for two reasons:
(a) as shown in Fig.~\ref{fig:memory-breakdown}, vision tokens greatly outnumber text prompt tokens;
(b) preliminary studies showed that reducing text tokens causes severe performance degradation.

\subsection{LightKV}
\label{sec:method-overview}

As illustrated in Fig.~\ref{fig:method-overview}, the pipeline of \ourmethod\ functions as follows:
At each specified decoder layer during the prefill stage, given a sequence of vision tokens, we first reconstruct their grid structure as in the original image.
These tokens are then partitioned into $w \times w$ small, non-overlapping windows, each containing an equal number of tokens.
Within each window, we perform message passing to compress vision tokens, simultaneously reducing KV size and the length of the vision input to the next decoder layer (see Sec.~\ref{sec:intra-window-compression}).
This is repeated in later layers with larger effective windows to achieve inter-window compression (see Sec.~\ref{sec:inter-window-compression}).

\subsubsection{Intra-window token compression}
\label{sec:intra-window-compression}

To address redundancy in vision tokens, we utilize message passing to aggregate information among tokens with low feature divergence (FD) (see Eq.~\ref{eq:similarity}), and subsequently eliminate redundant nodes within each window $\omega$.
The message passing and update procedure is applied independently to each window.
For notational clarity, we omit the subscript $\omega$ and use $v$ to denote the number of tokens in a window in Sec.~\ref{sec:intra-window-compression}.

\paragraph{Graph construction}
We map the vision tokens within each window to a bipartite graph.
For notational simplicity, we slightly abuse notation and use $\mathbf{x}$ to denote the embedding of a vision node.
\textbf{Step 1}: In each window, we first map each token $\mathbf{x}$ to a graph node, with $\mathcal{X} = \left\{\mathbf{x} | \mathbf{x} \in X_\mathtt{v} \right\}$.
Next, we partition the set of nodes into two near-equal subsets, $\mathcal{X}_{\mathcal{A}}$ and $\mathcal{X}_{\mathcal{B}}$ (shown in \textcolor{blue!60}{blue} and \textcolor{orange!70}{orange}, respectively, in Fig.~\ref{fig:method-overview}), by assigning tokens in an alternating manner: odd-indexed tokens to $\mathcal{X}_{\mathcal{A}}$ and even-indexed tokens to $\mathcal{X}_{\mathcal{B}}$.
We then construct a bipartite graph between the two subsets with edges $\mathcal{E}$:
\begin{equation}
    \mathcal{E}
        = \mathcal{X}_{\mathcal{A}} \times \mathcal{X}_{\mathcal{B}}
        = \big\{ (\mathbf{x}_{\alpha}, \mathbf{x}_{\beta}) \mid\ \forall\ \mathbf{x}_{\alpha} \in \mathcal{X}_{\mathcal{A}},\ \forall\ \mathbf{x}_{\beta} \in \mathcal{X}_{\mathcal{B}} \big\} ,
\end{equation}
where $\times$ denotes set cross product.
We modify the FD in~\citep{tranAcceleratingTransformersSpectrumPreserving2024, wangUnderstandingMitigatingMiscalibration2024} to weight each edge in the graph:
\begin{equation}
\label{eq:similarity}
    \operatorname{FD}(\alpha, \beta) = 1 - \frac{\langle \mathbf{x}_{\alpha}, \mathbf{x}_{\beta} \rangle}{||\mathbf{x}_{\alpha}||\ ||\mathbf{x}_{\beta}||},
\end{equation}
where $\langle \cdot, \cdot \rangle$ denotes the inner product and $|| \cdot ||$ is the $L^2$-norm.
\textbf{Step 2}: We compute the feature divergence $\operatorname{FD}(\alpha, \beta)$ for all bipartite pairings between $\mathcal{X}_{\mathcal{A}}$ and $\mathcal{X}_{\mathcal{B}}$.
These pairs are subsequently ranked in ascending order, and we construct the candidate set $\mathcal{T}_{\rho}$ by selecting the $\lfloor\rho v/2\rfloor$ pairs with the lowest $\operatorname{FD}$ values, where $\rho$ denotes the ratio of tokens removed.
Note that one-to-one matching is not enforced in $\mathcal{T}_{\rho}$: multiple nodes in $\mathcal{X}_\mathcal{A}$ may connect to the same node in $\mathcal{X}_\mathcal{B}$.
We then define the adjacency matrix $M \in \{0,1\}^{|\mathcal{X}_{\mathcal{A}}| \times |\mathcal{X}_{\mathcal{B}}|}$ as
\begin{equation}
    M_{\alpha,\beta} =
    \begin{cases}
        1, & \text{if } (\alpha,\beta) \in \mathcal{T}_{\rho}, \\
        0, & \text{otherwise}.
    \end{cases}
\end{equation}
Edges not in $\mathcal{T}_\rho$ are temporarily removed and unconnected nodes $\mathcal{X}_{\mathcal{R}} = \{ \mathbf{x}_{r} |\ \nexists\ \beta\ \text{s.t.} (r,\beta) \in \mathcal{T}_{\rho} \}$ are unchanged.

\paragraph{Token message passing}
In LVLMs, the heterogeneity of tokens introduces a challenge in evaluating the importance of each vision token, and prior works often disregard this by compressing tokens uniformly without accounting for their relative significance.
Instead, \ourmethod\ reuses the attention weights from the LLM decoder to estimate token importance, which are \textit{readily available during prefill without additional computation}, as shown in Eq.~\ref{eq:attention-matrix}.
This serves as a signal to preserve the visual features most relevant to the prompt, as measured by how strongly each vision token attends to the prompt tokens, and is used as guidance in the message-aggregation process.
\textbf{Step 3}: Given the $H$-headed attention matrix $A \in \mathbb{R}^{H \times (p+v) \times (p+v)}$, for a vision token with index $i$, we accumulate the attention of each vision token towards the prompt tokens:
\begin{equation}
    \label{eq:attention-factor}
    \xi_{i} = \sum_{h = 1}^{H} \ \sum_{j \in \mathcal{J}} \mathbf{A}[h, i, j],
\end{equation}
where $\mathcal{J}$ is the set of indices for the $p$ prompt tokens.
Here, $\mathbf{A}[h,i,j]$ denotes the attention weight where the query corresponds to vision token $i$ and the key corresponds to prompt token $j$.
Thus, $\xi_{i}$ captures how strongly each vision token aligns with the prompt semantics.
Next, we gather the attention for each window $\omega$ into vectors $\boldsymbol{\xi}_{\mathcal{A}} \in \mathbb{R}^{|\mathcal{X}_{\mathcal{A}}|}$ and $\boldsymbol{\xi}_{\mathcal{B}} \in \mathbb{R}^{|\mathcal{X}_{\mathcal{B}}|}$ with the same partitions as $\mathcal{X}_{\mathcal{A}}$ and $\mathcal{X}_{\mathcal{B}}$.
We update $X_{\mathcal{B}}$ by accumulating messages from its adjacent tokens:
\begin{equation}
    \label{eq:message-passing}
    X_{\mathcal{B}} =
        \underbrace{
            \big(\boldsymbol{\xi}_{\mathcal{B}} + M^{\top} \boldsymbol{\xi}_{\mathcal{A}} \big)^{-1} 
        }_{\text{(3) Normalize by sum of attentions}} \times \nonumber \Big(
            \underbrace{
                X_{\mathcal{B}} \odot \boldsymbol{\xi}_{\mathcal{B}}
            }_{\text{(1) Prompt-guidance for }\mathcal{B}}
            + \underbrace{
                M^{\top}
                \underbrace{
                    \left(\ X_{\mathcal{A}} \odot \boldsymbol{\xi}_{\mathcal{A}}\ \right)
                }_{\text{(1) Prompt-guidance for }\mathcal{A}}
            }_{\text{(2) Message passing over edges }M}
        \Big),
\end{equation}
where $(\cdot)^{-1}$ denotes element-wise inverse and $\odot$ is the Hadamard product.
This can be broken down into three parts:
\textbf{(1)} Messages from each token $\mathbf{x}_i$ are weighted by its attention $\xi_i$.
\textbf{(2)} Next, messages from the tokens in $\mathcal{X}_{\mathcal{A}}$ are passed to those in $\mathcal{X}_{\mathcal{B}}$ through the edges defined in $M$, updating tokens in $\mathcal{X}_{\mathcal{B}}$.
The choice of direction is arbitrary, and the reverse direction can be defined analogously.
\textbf{(3)} Finally, tokens in $\mathcal{X}_{\mathcal{B}}$ are normalized to remain scale-invariant.

Importantly, our aggregation operation utilizes the attention $\xi$ as guidance, ensuring the preservation of visual information that is most relevant to the prompt and the generation of the final response.
\textbf{Step 4}: After the update, the now-redundant nodes in $\mathcal{X}_\mathcal{A} \setminus \mathcal{X}_\mathcal{R}$ are deleted.
\textbf{Step 5}: Finally, the unchanged tokens $\mathcal{X}_{\mathcal{R}}$ and the updated $\mathcal{X}_{\mathcal{B}}$ are concatenated to form the final sequence of tokens for window $\omega$.

\paragraph{Complexity}
In contrast to computing fully pairwise FD among $v$ vision tokens in each window (which requires $\frac{1}{2} v (v-1)$ computations), the bipartite strategy reduces this by half to $\sim \frac{1}{4} v^2$.
We further validate this lower cost empirically in Table~\ref{tab:ablation-bipartite_vs_pairwise_flops}.

\paragraph{Difference from ToMe}
\ourmethod\ adopts a bipartite matching approach, similar to ToMe~\citep{bolyaTokenMergingYour2023}, to reduce the cost of pairwise calculations.
However, ToMe and subsequent methods assume all tokens are equally important, merging them without differentiation.
In contrast, \ourmethod\ uses cross-modality attention to guide message passing and aggregation, preserving the most relevant information during compression, yielding superior results (see Sec.~\ref{sec:experiments}).

\begin{wrapfigure}{r}{0.49\textwidth}
    \vspace{-2em}
    \centering
    \includegraphics[width=\linewidth]{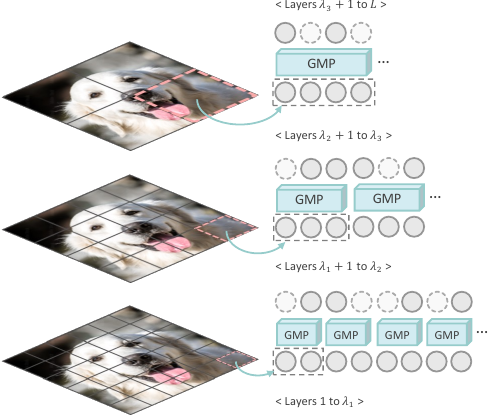}
    \caption{After each compression step, $w$ is reduced to allow message passing across greater spatial distances.}
    \label{fig:hierarchical-structure}
    \vspace{-3em}
\end{wrapfigure}

\subsubsection{Inter-window token compression} 
\label{sec:inter-window-compression}
In this section, the subscript $\omega$ is used to denote variables specific to an individual spatial window.

\paragraph{Window partitioning}
As discussed above, we split the entire set of vision tokens into window partitions in a non-overlapping manner.
Specifically, each window $\omega$ contains $v_\omega = v / (w \times w)$ vision tokens.
This reduces the total number of operations involved in computing FD measures from the original $\frac{1}{2} v (v-1)$ to $\frac{1}{2} \frac{v}{w^2} (\frac{v}{w^2}-1) \times w^2 \rightarrow \frac{1}{2} v (\frac{v}{w^2}-1)$.
Moreover, since spatially adjacent patches typically share semantic similarities, our window-based method confines message aggregation to within a small locality, preserving the positional information of tokens in the original image~\citep{songHierarchicalContextMerging2024, norouziALGMAdaptiveLocalthenGlobal2024}.
A global message passing strategy might inadvertently aggregate information from tokens representing unrelated entities, compromising locality and semantic coherence~\citep{xuGroupViTSemanticSegmentation2022, panLessMorePay2022}.

\paragraph{Hierarchical structure}
We adopt a \textit{hierarchical} compression strategy to improve efficiency, inspired by Swin-Transformer~\citep{liuSwinTransformerHierarchical2021}.  
Prior studies have shown that LLMs and LVLMs exhibit a layer-wise semantic hierarchy, where earlier layers tend to capture more local semantics while later layers progressively encode more global relations~\citep{duHowGPTLearns2025, liSemanticRoutingExploring2026}. 
Motivated by this observation, we design an iterative vision-token compression strategy that combines intra-window compression in earlier stages with progressively broader inter-window information aggregation in later stages.
Given an LVLM with $L$ layers, we perform $s$ compression iterations (where $s\!<\!L$), governed by three scheduling hyperparameters:
\begin{equation*}
\Lambda = [\lambda_1, \ldots, \lambda_s], \quad
\mathcal{W} = [w_1, \ldots, w_s], \quad
\mathcal{P} = [\rho_1, \ldots, \rho_s].
\end{equation*}
The hyperparameters $\Lambda$, $\mathcal{W}$, and $\mathcal{P}$ define the layer, window partition, and compression schedules, respectively.
For the $i$-th iteration, $\lambda_i$ denotes the target decoder layer, $w_i^2$ represents the number of window partitions, and $\rho_i$ specifies the per-step ratio in token reduction.
Specifically, the vision tokens exiting the decoder layer $\lambda_i$ are partitioned into $w_i^2$ partitions.
Within each window, vision tokens are compressed such that only a fraction $(1-\rho_i)$ remains for subsequent layers.
By enforcing $w_i > w_{i+1}$, we progressively expand the spatial scope of message passing across iterations, thus achieving hierarchical compression shown in Fig.~\ref{fig:hierarchical-structure}.

\subsection{Complexity analysis}
\label{sec:complexity}
Without any compression, the prefill stage processes in total $v \times L$ vision tokens.\footnote{We omit the double estimation of key and value cache for simplicity.}
With $s$ compression steps, the number of vision tokens processed during prefill now reduces to:
\begin{equation}
    \label{eq:complexity}
    v \times \ 
        \Bigg( \ 
            \underbrace{
                \lambda_{1} \vphantom{\prod_{j=1}^{i-1} (1 - \rho_{j})}
            }_{\text{(1)}}
            + 
            \underbrace{
                \sum_{i=2}^{s} \Big( \left(\lambda_{i} - \lambda_{i-1} \right) \prod_{j=1}^{i-1} (1 - \rho_{j}) \Big)
            }_{\text{(2)}} \\
            +
            \underbrace{
                \left(L - \lambda_{s} \right) \prod_{j=1}^{s} (1 - \rho_{j})
            }_{(3)} \ 
        \Bigg).
\end{equation}
If we consider the number of vision tokens in each layer independently, then the total number of vision tokens processed in $L$ decoder layers in a vanilla LVLM is simply $v \times L$.
However, the number of vision tokens reduces at every compression layer $\lambda_i$ (note that message passing and accumulation occur after each decoder layer $\lambda_i$).
$v \times \prod_{j=1}^{i-1}(1-\rho_j)$ denotes the number of remaining vision tokens after $i-1$ accumulation steps.
Then, between each pair of accumulation layers $\lambda_{i-1}$ and $\lambda_{i}$, the number of vision tokens processed is $\left(v \times \prod_{j=1}^{i-1}(1-\rho_j)\right) \times (\lambda_{i} - \lambda_{i-1})$.
Therefore, Eq.~\ref{eq:complexity} can be broken down into:
(1) number of vision tokens processed before the first accumulation step,
(2) number of vision tokens processed between the first and the last accumulation step, and
(3) number of vision tokens processed after the last accumulation step.
For example, in an LVLM with $L=40$ decoder layers, choosing  $\Lambda=[10,20,30]$ and $\mathcal{P}=[0.5,0.5,0.5]$ reduces the vision token count to $46.9\%$ of the baseline.

%% file: sec/4_experiments.tex
\section{Experiments}
\label{sec:experiments}

\begin{table*}[t!]
    \centering
    \caption{Results of \ourmethod\ on LLaVA models at $55\%$ vision token retention in the KV cache. \textbf{Avg \%} denotes the average of all performance metrics normalized against the Vanilla model. Methods are grouped by category and sorted by average score. ``NC'' and ``VW'' denote NoCaps and VizWiz, respectively.}
    \label{tab:main-results-llava}
    \vspace{0.5em}
    \begin{adjustbox}{max width=\linewidth}
        \input{tables/llava}

    \end{adjustbox}
    \vspace{-1em}
\end{table*}

\begin{table*}[!th]
    \centering
    \caption{Results of \ourmethod\ on InternVL2-8B at two vision token retention rates in KV cache. ``\textbf{Avg \%}'' denotes the average of all metrics normalized against the Vanilla model. Methods are sorted by average score. ``VW'' denotes VizWiz.}
    \label{tab:main-results-intern}
    \vspace{0.5em}
    \begin{adjustbox}{max width=\linewidth}
        \input{tables/internvl}

    \end{adjustbox}
    \vspace{-1em}
\end{table*}


\subsection{Experimental settings}

\paragraph{LVLM base models}
We evaluated the efficiency and performance of \ourmethod\ by applying it to eight open-source LVLMs: LLaVA-v1.5-13B, LLaVA-v1.5-7B, LLaVA-NeXT-13B, LLaVA-NeXT-7B, InternVL2-8B, EVE-7B-v1, EVE-7B-v1-HD, and Qwen2.5-VL-7B-Instruct.
LLaVA-v1.5 encodes 576 vision tokens per image, while LLaVA-NeXT uses 2,144.
In contrast, InternVL2 and Qwen2.5-VL adopt dynamic vision encoding, with token counts determined by image resolution.
It is worth noting that, unlike other models, which employ a dedicated image encoder, EVE is vision encoder-free.
These base models are labeled as \textit{Vanilla}.

\paragraph{Datasets}
We utilized eight publicly available large-scale benchmark datasets for evaluation:
Coco Caption~\citep{linMicrosoftCOCOCommon2014},
GQA~\citep{hudsonGQANewDataset2019}, 
MME~\citep{fuMMEComprehensiveEvaluation2024},
NoCaps (labeled ``NC'')~\citep{agrawalNocapsNovelObject2019},
Pope~\citep{liEvaluatingObjectHallucination2023}, 
SeedBench (``Seed'')~\citep{liSEEDBenchBenchmarkingMultimodal2024}, 
ScienceQA (``SQA'')~\citep{luLearnExplainMultimodal2022}, and 
VizWiz (``VW'')~\citep{gurariVizWizGrandChallenge2018}.
These benchmarks cover a wide range of tasks, from general, everyday image understanding to fine-grained image reasoning.
MME, Pope, SeedBench, and ScienceQA are limited to single-choice answers, while Coco Caption, GQA, NoCaps, and VizWiz involve open-ended responses comprising long sentences.

\paragraph{Compared baselines}
We adapted two existing techniques from other related domains: ToMe~\citep{bolyaTokenMergingYour2023} (labeled ``\textit{ToMe~(C)}'') and \textit{ElasticCache}~\citep{liuEfficientInferenceVision2024}.
For comparison, we implemented two random-eviction baselines: \textit{Rand} and \textit{ImgRand}.
\textit{Rand} and \textit{ElasticCache} prune both text and vision tokens, whereas \textit{ImgRand} and ToMe reduce vision tokens only.
It is important to note that the previously mentioned methods perform token reduction \textit{after} the prefill stage.
Additionally, for token reduction \textit{during} prefill, we implemented ToMe (labeled ``\textit{ToMe~(P)}'') and four recent SOTA strategies: \textit{FastV}~\citep{chenImageWorth12024}, \textit{PiToMe}~\citep{tranAcceleratingTransformersSpectrumPreserving2024}, \textit{ToFu}~\citep{kimTokenFusionBridging2024} and \textit{HiRED}~\citep{arifHiREDAttentionguidedToken2025}.\footnote{HiRED uses the same model but with HuggingFace optimizations; efficiency metrics are omitted for fairness.}

\paragraph{Implementation details}
In our experiments, we retain the default parameters of the LVLM backbones and use greedy decoding for reproducibility.
For FastV, we adopt the reported optimal setting of $K=2$ and vary only $R$ to control the KV cache pruning ratio.
For other methods, we adapted them to work with the LVLM backbones as faithfully as possible.
To ensure consistency, we fix the schedule of \ourmethod's compression layers $\Lambda$, compression ratios $\mathcal{P}$, and window sizes $\mathcal{W}$ across all benchmarks for each LVLM model.
We utilized \textit{lmms-eval}~\citep{zhangLMMsEvalRealityCheck2025} for all benchmark evaluations.
We profiled the time-to-first-token (TTFT) and the generation latency for 100 tokens by averaging over 10 runs on an NVIDIA A100 GPU.

\begin{figure}[!t]
    \centering
    \begin{minipage}{0.60\textwidth}
        \centering
        \captionof{table}{Results of \ourmethod\ on EVE-7B-v1 models at 55\% retention of vision tokens in the KV cache. ``NC'' and ``VW'' denote NoCaps and VizWiz, respectively.}
        \label{tab:main-results-eve}
        \vspace{0.2em}
        \begin{adjustbox}{max width=\linewidth}
            \input{tables/evev1}
        \end{adjustbox}
    \end{minipage}
    \hfill
    \begin{minipage}{0.38\textwidth}
        \centering
        \includegraphics[width=\linewidth]{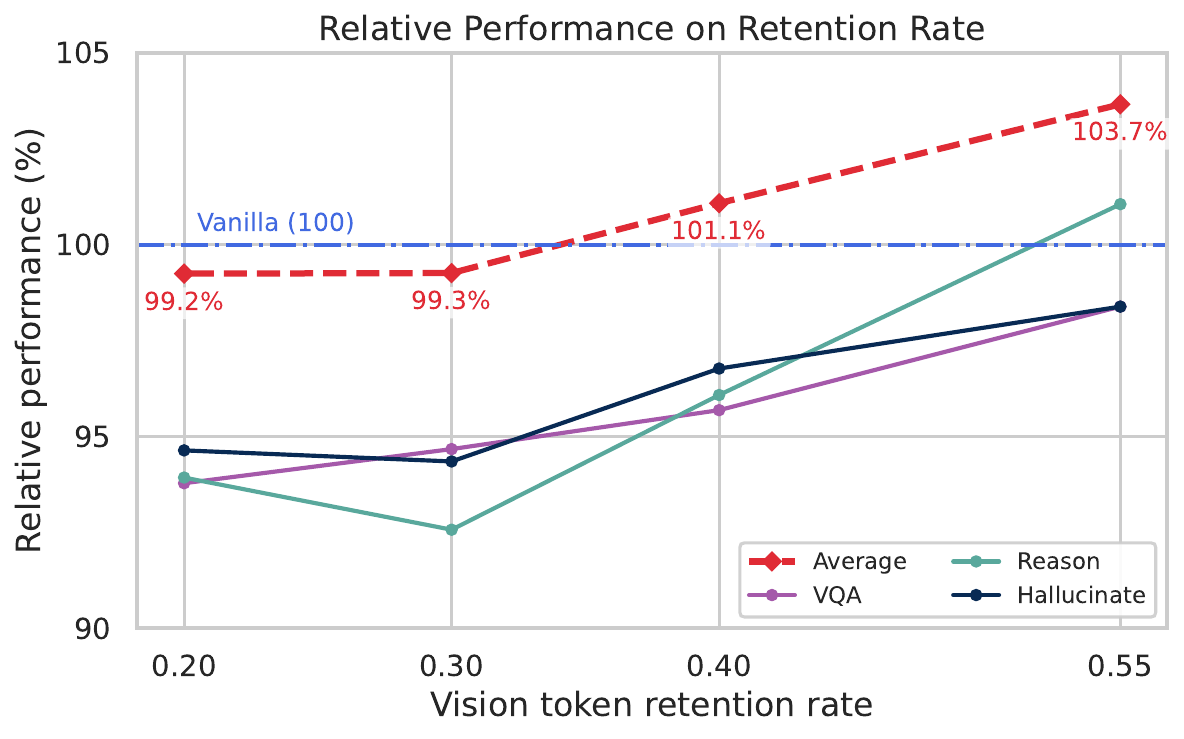}
        \caption{
            Effect of varying retention rates on Qwen2.5-VL. The ``Average'' curve summarizes the overall performance trend across Reasoning, VQA, Hallucination and Captioning.\protect\footnotemark
        }
        \label{fig:qwen-compression}
    \end{minipage}
\end{figure}
\footnotetext{The Captioning trend is omitted because its performance remains above 105\%, exceeding the current vertical axis range.}

\subsection{Main results}
\label{sec:overall-results}

We compare the performance of \ourmethod\ with other SOTA methods on LLaVA models (Table~\ref{tab:main-results-llava}), InternVL (Table~\ref{tab:main-results-intern}), EVE (Table~\ref{tab:main-results-eve}) and Qwen2.5-VL (Fig.~\ref{fig:qwen-compression} and Table~\ref{tab:main-results-qwen} in the appendix).
For each LVLM model, we selected the optimal configurations of $\Lambda$ and $\mathcal{W}$ based on performance on Coco and MME, and applied these hyperparameters to the remaining benchmarks.
We also profiled efficiency metrics, including FLOPs, KV cache memory (from prompt, vision, and generated tokens), and time to first token (TTFT) when generating 100 tokens (standard deviation reported in the supplementary).
Our key findings are summarized as follows:
\begin{itemize}[leftmargin=1em]
    \item Tables~\ref{tab:main-results-llava}, \ref{tab:main-results-intern}, \ref{tab:main-results-eve} and~\ref{tab:main-results-qwen} show that \ourmethod\ consistently preserves the performance of the base LVLMs across most benchmarks. In some cases, our method surpasses the performance of LVLMs without compression.

    \item Compared to methods applied \textit{during the prefill stage} (see Table~\ref{tab:main-results-llava}), \ourmethod\ either outperforms or achieves highly competitive results, ranking first in 3 out of 4 LLaVA models and second in the remaining one. Furthermore, baseline methods often obtain lower FLOPs or memory at the cost of larger performance degradation, whereas LightKV provides a stronger performance-efficiency tradeoff.

    \item Our method yields the most consistent performance across the models, while others exhibit inconsistent rankings due to substantial degradations. For example, FastV performs well on LLaVA-v1.5 models, but shows substantial drops on LLaVA-NeXT models. We attribute this to its pruning strategy, which removes vision tokens solely based on early-layer visual attention scores. Given that LLaVA-v1.5 encodes only 576 vision tokens while LLaVA-NeXT processes 2,144, early-layer attention in the latter is far sparser and less reliable as an importance signal, causing FastV to prematurely discard tokens that later contribute to cross-modal reasoning, a shortfall mitigated by our hierarchical strategy.

    \item At even more aggressive compression ratios (\eg, retaining 20\% and 30\%), \ourmethod\ is capable of retaining 99\% average performance across multiple benchmarks on Qwen2.5-VL (Fig.~\ref{fig:qwen-compression} and Table~\ref{tab:main-results-qwen} in the appendix), further highlighting its robustness.

    \item \ourmethod\ is compatible not only with vision encoder-based LVLMs, but also with encoder-free models such as EVE, which seek to reduce the strong inductive bias in the vision encoders. As shown in Table~\ref{tab:main-results-eve}, our approach substantially outperforms FastV at the same compression rate, and is better at preserving the original capabilities of the LVLMs.

    \item Post-prefill approaches substantially degrade performance on open-ended tasks, \eg, Coco and NoCaps. Additionally, they yield minimal improvements in efficiency, since the prefill stage remains the dominant memory and latency bottleneck. In contrast, \ourmethod\ operates during prefill within the decoder layers, resulting in significantly lower compute cost and memory footprint while achieving stronger performance.
    
\end{itemize}

\begin{table}[!t]
    \centering
    \caption{Comparison of prompt-guided weighting to uniform and random variants at 55\% vision token retention.}
    \label{tab:ablation-prompt-guidance}
    \begin{adjustbox}{width=0.7\linewidth}
        \input{tables/ablation_prompt_guidance}
    \end{adjustbox}
\end{table}

\subsection{Additional experiments}

\paragraph{Effect of prompt guidance}
To isolate the specific contribution of prompt-aware guidance within \ourmethod, we conduct an ablation study comparing our approach against variants utilizing uniform and random attention weights. 
As reported in Table~\ref{tab:ablation-prompt-guidance}, substituting our prompt-guided mechanism with these simpler weighting schemes results in consistent performance degradation across benchmarks. 
These findings validate the performance gains stemming from the cross-modal signals during compression.

\paragraph{Latency profiling}
Table~\ref{tab:latency} illustrates the reduction in TTFT and generation latency over 100 tokens achieved by \ourmethod.
As our approach requires explicit attention matrices, it is incompatible with I/O-optimized mechanisms like FlashAttention~\citep{daoFlashAttentionFastMemoryEfficient2022}.
To overcome this, we selectively switch to eager computation in the small subset ($s \ll L$) of layers where compression is applied, while retaining the optimized attention implementation for the majority.
The marginal overhead is offset by the increased throughput achieved by processing fewer vision tokens in the downstream layers.
See Sec.~\ref{sec:supp-latency} for more details.

\begin{table}[!t]
    \centering
    \caption{TTFT (ms) and 100-token generation latency (s) $\pm$ Std.\;Dev. on LLaVA 13B models.}
    \label{tab:latency}
    \begin{adjustbox}{max width=0.7\linewidth}
        \input{tables/latency}
    \end{adjustbox}
    \vspace{-0.5em}
\end{table}

\paragraph{Influence of hierarchical compression}
We conducted experiments at the same compression layer $\lambda$ while varying $\mathcal{W}$, as presented in Table~\ref{tab:window-ablation}.
Across different compression layers $\lambda$, the results show a similar general trend: there is more pronounced degradation with a global compression strategy $w=1$, likely due to the inadvertent destruction of spatial locality~\citep{xuGroupViTSemanticSegmentation2022, panLessMorePay2022, songHierarchicalContextMerging2024, norouziALGMAdaptiveLocalthenGlobal2024}.

We further evaluate the relative merits of our hierarchical compression in Table~\ref{tab:ablation-window-inter-vs-intra}.
We conduct additional experiments that perform compression directly on the full set of vision tokens (global-only), and only within fixed windows (local-only).
Our results demonstrate that both variants underperform when compared to our strategy.
This suggests that the efficacy of \ourmethod\ stems from the progressive expansion of the compression scope across stages, which balances local feature preservation with integration of global semantics.

Lastly, we summarize the FLOPs and KV cache memory usage for different inference configurations in Table~\ref{tab:profile-ablation}, which shows that changing $\mathcal{W}$ has limited impact on aggregated FLOPs and memory under the same compression schedule.

\begin{table}[!t]
    \centering
    \caption{Effect of varying window sizes $w$ at different compression layers on the performance of InternVL-8B across benchmarks. ``VW'' denotes VizWiz.}
    \label{tab:window-ablation}
    \begin{adjustbox}{width=0.7\linewidth}
        \input{tables/ablation_window}
    \end{adjustbox}
    \vspace{-0.5em}
\end{table}

\begin{table}[!th]
    \centering
    \caption{Performance comparison to global-only and local-only compression at $55\%$ vision token retention.}
    \label{tab:ablation-window-inter-vs-intra}
    \begin{adjustbox}{width=0.7\linewidth}
        \input{tables/ablation_window_inter_vs_intra}
    \end{adjustbox}
\end{table}

\begin{table}[!th]
    \centering
    \caption{Profiling results by varying compression layers $\Lambda$ and window sizes $\mathcal{W}$ on LLaVA 13B models.}
    \label{tab:profile-ablation}
    \begin{adjustbox}{width=0.7\linewidth}
        \input{tables/profile_ablation}

    \end{adjustbox}
    \vspace{-0.5em}
\end{table}

\paragraph{Influence of compression layers}
We investigate the impact of varying layers for token compression, as illustrated in Figure~\ref{fig:layer-ablation} in the appendix.
Trends between the compression layer and model performance reveal that compressing in the shallow layers has a more substantial impact on performance.
This effect is particularly pronounced in VizWiz, where LVLMs must refrain from answering (\eg, when the ground truth is ``unanswerable'').
Compression in the deeper layers yields performance nearly identical to the base LVLM models, but offers little reduction in memory usage.

Additional ablation studies, including bipartite vs.\! full pairwise matching for computing FD (Tables~\ref{tab:ablation-bipartite_vs_pairwise} and~\ref{tab:ablation-bipartite_vs_pairwise_flops}), similarity metrics (Table~\ref{tab:ablation-similarity-metric}), and FastV under hyperparameter tuning (Table~\ref{tab:fastv_k_sweep}) are provided in Appendix Sec.~\ref{sec:app-ablation}.

%% file: tables/llava.tex
\begin{tabular}{cl|rrr|cccccccc|r}
\toprule
& \multirow{2}{*}{Method} &
  FLOPs &
  Mem &
  TTFT &
  \multirow{2}{*}{Coco} &
  \multicolumn{2}{c}{MME} &
  \multirow{2}{*}{NC} &
  \multicolumn{2}{c}{Pope} &
  \multirow{2}{*}{Seed} &
  \multirow{2}{*}{VW} &
  \multirow{2}{*}{\textbf{Avg \%}} \\
\cmidrule(l){7-8} \cmidrule(l){10-11}
& & \multicolumn{1}{c}{\textit{Tera} $\downarrow$} & \multicolumn{1}{c}{\textit{GB} $\downarrow$} & \multicolumn{1}{c}{\textit{sec} $\downarrow$} & & \multicolumn{1}{c}{C} & \multicolumn{1}{c}{P} & & \multicolumn{1}{c}{Acc} & \multicolumn{1}{c}{F1} & & & \\
\midrule

\rowcolor{gray!15} \cellcolor{white} \multirow{12}{*}{\rotatebox{90}{\textbf{LLaVA-v1.5-13B}}} & Vanilla & 19.4 & 0.55 & 0.130 & 1.16 & 295.4 & 1532.0 & 1.09 & 0.87 & 0.86 & 0.69 & 0.57 & 100.00 \\
& \multicolumn{12}{c}{\textbf{Post prefill}} \\
& Elastic  & 19.3 & 0.31 & 0.598 & 0.96 & 295.4 & 1534.5 & 0.87 & 0.43 & 0.96 & OOM  & 0.14 & 68.54 \\
& Rand     & 19.0 & 0.31 & 0.134 & 0.48 & 295.4 & 1532.9 & 0.46 & 0.46 & 0.89 & 0.70 & 0.13 & 70.53 \\
& ImgRand  & 19.0 & 0.31 & 0.134 & 0.95 & 295.4 & 1532.9 & 0.86 & 0.69 & 0.91 & 0.70 & 0.19 & 85.09 \\
& ToMe (C) & 19.0 & 0.33 & 0.141 & 1.00 & 295.4 & 1532.9 & 0.92 & 0.79 & 0.88 & 0.70 & 0.18 & 87.10 \\
& \multicolumn{12}{c}{\textbf{During prefill}} \\
& ToFu      & 12.6 & 0.37 & 0.094 & 1.14 & 292.1 & 1535.7 & 1.08 & 0.86 & 0.86 & 0.38 & 0.55 & 93.36 \\
& PiToMe    & 12.6 & 0.37 & 0.093 & 1.14 & 297.5 & 1529.0 & 1.07 & 0.87 & 0.85 & 0.38 & 0.55 & 93.42 \\
& ToMe (P)  & 12.6 & 0.37 & 0.094 & 1.16 & 297.5 & 1529.9 & 1.07 & 0.87 & 0.86 & 0.39 & 0.55 & 93.96 \\
\rowcolor{cyan!10} \cellcolor{white} & \ourmethod
            & 12.6 & 0.37 & 0.098 & 1.15 & 302.1 & 1543.8 & 1.08 & 0.87 & 0.86 & 0.69 & 0.56 & 99.94 \\
& FastV     & 12.4 & 0.36 & 0.085 & 1.16 & 308.9 & 1546.6 & 1.09 & 0.86 & 0.85 & 0.68 & 0.57 & \textbf{100.22} \\

\midrule

\rowcolor{gray!15} \cellcolor{white} \multirow{12}{*}{\rotatebox{90}{\textbf{LLaVA-v1.5-7B}}} & Vanilla & 10.2 & 0.35 & 0.078 & 1.10 & 355.7 & 1509.6 & 1.05 & 0.87 & 0.86 & 0.66 & 0.54 & 100.00 \\
& \multicolumn{12}{c}{\textbf{Post prefill}} \\
& Elastic  & 10.2 & 0.20 & 0.449 & 0.41 & 350.4 & 1508.9 & 0.30 & 0.30 & 0.93 & OOM & 0.09 & 52.95 \\
& Rand      & 9.9 & 0.21 & 0.081 & 0.13 & 350.4 & 1508.9 & 0.10 & 0.74 & 0.87 & 0.66 & 0.11 & 65.80 \\
& ToMe (C)  & 10.0 & 0.20 & 0.086 & 0.13 & 350.4 & 1508.9 & 0.09 & 0.87 & 0.86 & 0.66 & 0.18 & 69.02 \\
& ImgRand   & 9.9 & 0.20 & 0.082 & 0.22 & 350.4 & 1508.9 & 0.16 & 0.86 & 0.86 & 0.66 & 0.16 & 70.27 \\
& \multicolumn{12}{c}{\textbf{During prefill}} \\
& HiRED & - & - & - & 1.03 & 335.0 & 1452.0 & 1.00 & 0.85 & 0.83 & 0.66 & 0.53 & 96.45 \\
& ToMe (P)  & 6.6 & 0.23 & 0.058 & 1.09 & 319.6 & 1490.5 & 1.01 & 0.87 & 0.86 & 0.66 & 0.52 & 97.52 \\
& PiToMe    & 6.6 & 0.23 & 0.058 & 1.08 & 341.0 & 1498.5 & 1.02 & 0.86 & 0.85 & 0.65 & 0.51 & 97.63 \\
& ToFu      & 6.6 & 0.23 & 0.058 & 1.09 & 340.0 & 1482.3 & 1.02 & 0.86 & 0.85 & 0.66 & 0.52 & 97.98 \\
& FastV     & 5.3 & 0.22 & 0.052 & 1.10 & 351.1 & 1513.7 & 1.04 & 0.85 & 0.83 & 0.66 & 0.54 & 99.03 \\
\rowcolor{cyan!10} \cellcolor{white} & \ourmethod
            & 6.6 & 0.23 & 0.065 & 1.11 & 357.5 & 1519.8 & 1.03 & 0.87 & 0.86 & 0.66 & 0.53 & \textbf{99.79} \\

\midrule

\rowcolor{gray!15} \cellcolor{white} \multirow{12}{*}{\rotatebox{90}{\textbf{LLaVA-NeXT-13B}}} & Vanilla & 65.0 & 1.75 & 0.656 & 1.02 & 318.9 & 1575.1 & 0.88 & 0.88 & 0.86 & 0.69 & 0.64 & 100.00 \\
& \multicolumn{12}{c}{\textbf{Post prefill}} \\
& Elastic  & - & - & 2.302 & OOM & OOM & OOM & OOM & OOM & OOM & OOM & OOM & 0.00 \\
& Rand     & 60.8 & 0.91 & 0.651 & 0.06 & 318.9 & 1575.1 & 0.04 & 0.82 & 0.86 & 0.69 & 0.08 & 64.51 \\
& ToMe (C) & 61.3 & 0.93 & 0.683 & 0.07 & 318.9 & 1575.1 & 0.05 & 0.87 & 0.86 & 0.69 & 0.08 & 65.48 \\
& ImgRand  & 60.8 & 0.91 & 0.652 & 0.07 & 318.9 & 1575.1 & 0.05 & 0.87 & 0.86 & 0.69 & 0.08 & 65.50 \\
& \multicolumn{12}{c}{\textbf{During prefill}} \\
& ToMe (P)  & 37.3 & 1.05 & 0.394 & 0.97 & 308.5 & 1551.0 & 0.84 & 0.87 & 0.86 & 0.34 & 0.60 & 90.96 \\
& ToFu      & 37.3 & 1.05 & 0.394 & 0.97 & 305.0 & 1539.5 & 0.83 & 0.88 & 0.87 & 0.36 & 0.60 & 91.31 \\
& PiToMe    & 37.3 & 1.05 & 0.396 & 0.98 & 311.9 & 1558.2 & 0.86 & 0.87 & 0.86 & 0.34 & 0.60 & 91.56 \\
& FastV     & 36.1 & 1.04 & 0.321 & 0.91 & 311.1 & 1477.5 & 0.81 & 0.82 & 0.78 & 0.68 & 0.61 & 93.80 \\
\rowcolor{cyan!10} \cellcolor{white} & \ourmethod
            & 37.3 & 1.05 & 0.383 & 0.96 & 326.1 & 1576.5 & 0.83 & 0.87 & 0.86 & 0.69 & 0.61 & \textbf{98.12} \\

\midrule
\rowcolor{gray!15} \cellcolor{white} \multirow{12}{*}{\rotatebox{90}{\textbf{LLaVA-NeXT-7B}}} & Vanilla & 34.8 & 1.12 & 0.397 & 1.00 & 330.0 & 1528.2 & 0.88 & 0.88 & 0.86 & 0.68 & 0.61 & 100.00 \\
& \multicolumn{12}{c}{\textbf{Post prefill}} \\
& Elastic  & 34.7 & 0.58 & 1.693 & 0.02 & 332.1 & 1519.3 & 0.01 & 0.18 & 0.90 & OOM & 0.08 & 42.67 \\
& Rand     & 32.2 & 0.58 & 0.397 & 0.02 & 322.5 & 1523.2 & 0.01 & 0.65 & 0.87 & 0.68 & 0.08 & 61.08 \\
& ImgRand  & 32.2 & 0.58 & 0.397 & 0.02 & 322.5 & 1523.2 & 0.02 & 0.85 & 0.87 & 0.68 & 0.08 & 64.06 \\
& ToMe (C) & 32.5 & 0.60 & 0.416 & 0.03 & 322.5 & 1523.2 & 0.02 & 0.87 & 0.86 & 0.68 & 0.08 & 64.33 \\
& \multicolumn{12}{c}{\textbf{During prefill}} \\
& FastV     & 18.5 & 0.65 & 0.197 & 0.88 & 265.4 & 1341.3 & 0.78 & 0.81 & 0.77 & 0.69 & 0.58 & 90.37 \\
& HiRED     & - & - & -           & 0.73 & 297.9 & 1398.9 & 0.67 & 0.88 & 0.87 & 0.66 & 0.58 & 90.68 \\
& ToMe (P)  & 21.1 & 0.67 & 0.245 & 0.93 & 292.9 & 1419.0 & 0.78 & 0.88 & 0.87 & 0.65 & 0.57 & 94.18 \\
& ToFu      & 20.0 & 0.67 & 0.245 & 0.93 & 295.4 & 1427.2 & 0.78 & 0.88 & 0.87 & 0.66 & 0.57 & 94.52 \\
& PiToMe    & 20.0 & 0.67 & 0.247 & 0.94 & 292.1 & 1415.5 & 0.79 & 0.88 & 0.87 & 0.65 & 0.58 & 94.58 \\
\rowcolor{cyan!10} \cellcolor{white} & \ourmethod
            & 22.3 & 0.67 & 0.259 & 0.98 & 338.6 & 1517.3 & 0.83 & 0.88 & 0.86 & 0.69 & 0.58 & \textbf{98.85} \\
\bottomrule
\end{tabular}

%% file: tables/internvl.tex
\begin{tabular}{l|rrr|rrrrrrrr|r}
\toprule
\multirow{2}{*}{Method} & 
FLOPs & 
Mem & 
TTFT &
\multirow{2}{*}{Coco} & 
\multirow{2}{*}{GQA} & 
\multicolumn{2}{c}{MME} & 
\multicolumn{2}{c}{Pope} & 
\multirow{2}{*}{SQA} & 
\multirow{2}{*}{VW} &
\multirow{2}{*}{\textbf{Avg \%}} \\
\cmidrule(l){7-8} \cmidrule(l){9-10}
& \multicolumn{1}{c}{\textit{Tera} $\downarrow$} & \multicolumn{1}{c}{\textit{GB} $\downarrow$} & \multicolumn{1}{c}{\textit{sec} $\downarrow$} & & & \multicolumn{1}{c}{C} & \multicolumn{1}{c}{P} & \multicolumn{1}{c}{Acc} & \multicolumn{1}{c}{F1} & & & \\
\midrule
\rowcolor{gray!15} Vanilla & 35.7 & 0.24 & 0.460 & 0.90 & 0.63 & 587.5 & 1623.8 & 0.88 & 0.87 & 0.97 & 0.61 & 100.00 \\
\multicolumn{13}{c}{\textbf{During prefill, retain 60\% vision tokens}} \\
FastV       & 24.8 & 0.15 & 0.520 & 0.80 & 0.50 & 569.6 & 1610.9 & 0.47 & 0.87 & 0.49 & 0.53 & 81.90 \\
ToFu        & 22.1 & 0.15 & 0.395 & 0.81 & 0.62 & 502.1 & 1575.5 & 0.87 & 0.86 & 0.94 & 0.60 & 95.49 \\
PiToMe      & 22.1 & 0.15 & 0.396 & 0.99 & 0.60 & 461.8 & 1545.3 & 0.87 & 0.86 & 0.90 & 0.60 & 95.99 \\
ToMe (P)    & 22.1 & 0.15 & 0.397 & 0.87 & 0.62 & 551.4 & 1621.8 & 0.87 & 0.86 & 0.95 & 0.60 & 97.86 \\
\rowcolor{cyan!10} \ourmethod
            & 23.1 & 0.15 & 0.391 & 0.91 & 0.63 & 590.0 & 1623.8 & 0.88 & 0.87 & 0.97 & 0.61 & \textbf{100.19} \\
\midrule
\multicolumn{13}{c}{\textbf{During prefill, retain 55\% vision tokens}} \\
FastV       & 22.9 & 0.14 & 0.517 & 0.68 & 0.47 & 582.1 & 1611.1 & 0.56 & 0.85 & 0.46 & 0.48 & 79.49 \\
PiToMe      & 22.1 & 0.15 & 0.396 & 1.00 & 0.61 & 442.9 & 1575.5 & 0.87 & 0.86 & 0.90 & 0.57 & 95.54 \\
ToMe (P)    & 22.1 & 0.15 & 0.397 & 0.81 & 0.62 & 503.9 & 1570.0 & 0.87 & 0.86 & 0.95 & 0.60 & 95.62 \\
ToFu        & 22.1 & 0.15 & 0.395 & 0.75 & 0.62 & 541.8 & 1619.1 & 0.87 & 0.85 & 0.95 & 0.60 & 95.82 \\
\rowcolor{cyan!10} \ourmethod
            & 23.1 & 0.15 & 0.391 & 0.88 & 0.62 & 590.0 & 1623.8 & 0.88 & 0.87 & 0.97 & 0.61 & \textbf{99.58} \\ 

\bottomrule
\end{tabular}

%% file: tables/evev1.tex
\begin{tabular}{l|ccccccc|c}
\toprule
\multirow{2}{*}{Method} & 
  \multirow{2}{*}{Coco} & 
  \multicolumn{2}{c}{MME} & 
  \multirow{2}{*}{NC} & 
  \multicolumn{2}{c}{Pope} & 
  \multirow{2}{*}{VW} & 
  \multirow{2}{*}{\textbf{Avg \%}} \\ 
\cmidrule(l){3-4} \cmidrule(l){6-7}
& & \multicolumn{1}{c}{C} & \multicolumn{1}{c}{P} & & \multicolumn{1}{c}{Acc} & \multicolumn{1}{c}{F1} & \\
\midrule
\multicolumn{9}{c}{\textbf{EVE-7B-v1}} \\
\rowcolor{gray!15} Vanilla & 0.96 & 269.2 & 1230.8 & 0.94 & 0.84 & 0.83 & 0.46 & 100.00 \\
FastV     & 0.85 & 259.3 & 1144.5 & 0.78 & 0.80 & 0.77 & 0.44 & 92.07 \\
\rowcolor{cyan!10} LightKV   & 1.00 & 269.3 & 1203.1 & 0.93 & 0.84 & 0.83 & 0.43 & \textbf{99.20} \\
\midrule
\multicolumn{9}{c}{\textbf{EVE-7B-v1-HD}} \\
\rowcolor{gray!15} Vanilla & 1.05 & 304.6 & 1314.1 & 1.02 & 0.86 & 0.85 & 0.56 & 100.00 \\
FastV        & 0.97 & 290.3 & 1238.6 & 0.93 & 0.83 & 0.82 & 0.55 & 94.90 \\
\rowcolor{cyan!10} LightKV      & 0.97 & 291.4 & 1308.9 & 0.94 & 0.86 & 0.85 & 0.54 & \textbf{96.61} \\
\bottomrule
\end{tabular}

%% file: tables/ablation_prompt_guidance.tex
\begin{tabular}{l|cccccccc|c}
\toprule
\multirow{2}{*}{Method} &
\multirow{2}{*}{Coco} & 
\multicolumn{2}{c}{MME} & 
\multirow{2}{*}{NC} & 
\multicolumn{2}{c}{Pope} & 
\multirow{2}{*}{Seed} & 
\multirow{2}{*}{VW} &
\multirow{2}{*}{\textbf{Avg\ \%}} \\
\cmidrule(l){3-4} \cmidrule(l){6-7}
& & \multicolumn{1}{c}{C} & \multicolumn{1}{c}{P} & & \multicolumn{1}{c}{Acc} & \multicolumn{1}{c}{F1} & & & \\

\midrule
\multicolumn{10}{c}{\textbf{LLaVA-v1.5-13B}} \\
Uniform & 1.14 & 299.6 & 1535.0 & 1.06 & 0.87 & 0.86 & 0.39 & 0.55 & 93.78 \\
Random & 1.14 & 300.0 & 1534.5 & 1.07 & 0.87 & 0.85 & 0.39 & 0.56 & 93.96 \\
\rowcolor{cyan!10} Prompt & 1.15 & 302.1 & 1543.8 & 1.08 & 0.87 & 0.86 & 0.69 & 0.56 & \textbf{99.94} \\
\midrule
\multicolumn{10}{c}{\textbf{LLaVA-NeXT-13B}} \\
Uniform & 0.98 & 311.0 & 1547.3 & 0.85 & 0.86 & 0.85 & 0.35 & 0.60 & 91.18 \\
Random & 0.97 & 311.1 & 1542.3 & 0.84 & 0.86 & 0.86 & 0.34 & 0.59 & 90.65 \\
\rowcolor{cyan!10} Prompt & 0.96 & 326.1 & 1576.5 & 0.83 & 0.87 & 0.86 & 0.69 & 0.61 & \textbf{98.12} \\
\bottomrule
\end{tabular}

%% file: tables/latency.tex

\begin{tabular}{l|cc|cc}
\toprule
Method & TTFT (ms) & Gen latency (s) & TTFT (ms) & Gen latency (s) \\
\midrule
& \multicolumn{2}{c}{\textbf{LLaVA-v1.5-13B}} & \multicolumn{2}{c}{\textbf{LLaVA-NeXT-13B}} \\
Vanilla  & 130 $\pm$ 0.393 & 2.89 $\pm$ 0.004 & 656 $\pm$ 1.479 & 3.79 $\pm$ 0.013\\
LightKV  &  98 $\pm$ 1.045 & 2.80 $\pm$ 0.004 & 383 $\pm$ 0.878 & 3.29 $\pm$ 0.002 \\
\bottomrule
\end{tabular}

%% file: tables/ablation_window.tex
\begin{tabular}[t]{c|crrrrrrrr}
\toprule
\multirow{2}{*}{Method} &
\multirow{2}{*}{$\mathcal{W}$} &
\multirow{2}{*}{Coco} & 
\multirow{2}{*}{GQA} & 
\multicolumn{2}{c}{MME} & 
\multicolumn{2}{c}{Pope} & 
\multirow{2}{*}{SQA} & 
\multirow{2}{*}{VW} \\
\cmidrule(l){5-6} \cmidrule(l){7-8}
& & & & \multicolumn{1}{c}{C} & \multicolumn{1}{c}{P} & \multicolumn{1}{c}{Acc} & \multicolumn{1}{c}{F1} & & \\
\midrule
\rowcolor{gray!15}{Vanilla} & - & 0.90 & 0.63 & 587.5 & 1623.8 & 0.88 & 0.87 & 0.97 & 0.61 \\
\midrule
\multirow{8}{*}{\rotatebox{90}{\ourmethod}} 
& \multicolumn{9}{c}{$\boldsymbol{\lambda}\mathbf{=3}$} \\
& $1$ & 0.80 & 0.62 & 547.5 & 1602.5 & 0.87 & 0.86 & 0.95 & 0.60 \\
& $2$ & 0.83 & 0.59 & 555.0 & 1621.1 & 0.87 & 0.86 & 0.96 & 0.60 \\
& $4$ & 0.90 & 0.60 & 546.8 & 1594.8 & 0.87 & 0.85 & 0.95 & 0.60 \\
& \multicolumn{9}{c}{$\boldsymbol{\lambda}\mathbf{=14}$} \\
& $1$ & 0.89 & 0.62 & 577.1 & 1615.8 & 0.87 & 0.86 & 0.97 & 0.61 \\
& $2$ & 0.90 & 0.62 & 577.1 & 1620.3 & 0.87 & 0.86 & 0.97 & 0.61 \\
& $4$ & 0.92 & 0.62 & 577.9 & 1617.5 & 0.88 & 0.86 & 0.97 & 0.61 \\
\bottomrule
\end{tabular}

%% file: tables/ablation_window_inter_vs_intra.tex
\begin{tabular}{l|cccccccc|c}
\toprule
\multirow{2}{*}{Strategy} &
\multirow{2}{*}{Coco} & 
\multicolumn{2}{c}{MME} & 
\multirow{2}{*}{NC} & 
\multicolumn{2}{c}{Pope} & 
\multirow{2}{*}{Seed} & 
\multirow{2}{*}{VW} &
\multirow{2}{*}{\textbf{Avg\ \%}} \\
\cmidrule(l){3-4} \cmidrule(l){6-7}
& & \multicolumn{1}{c}{C} & \multicolumn{1}{c}{P} & & \multicolumn{1}{c}{Acc} & \multicolumn{1}{c}{F1} & & & \\

\midrule
\multicolumn{10}{c}{\textbf{LLaVA-v1.5-13B}} \\
Global-only & 1.15 & 299.6 & 1530.2 & 1.08 & 0.87 & 0.85 & 0.39 & 0.55 & 93.92 \\
Local-only  & 1.15 & 290.0 & 1529.9 & 1.08 & 0.87 & 0.85 & 0.38 & 0.56 & 93.55 \\
\rowcolor{cyan!10} Ours  & 1.15 & 302.1 & 1543.8 & 1.08 & 0.87 & 0.86 & 0.69 & 0.56 & \textbf{99.94} \\
\hline
\multicolumn{10}{c}{\textbf{LLaVA-NeXT-13B}} \\
Global-only & 0.98 & 311.1 & 1549.8 & 0.85 & 0.87 & 0.86 & 0.34 & 0.60 & 91.31 \\
Local-only  & 0.97 & 318.5 & 1543.8 & 0.85 & 0.87 & 0.86 & 0.34 & 0.60 & 91.44 \\
\rowcolor{cyan!10} Ours  & 0.96 & 326.1 & 1576.5 & 0.83 & 0.87 & 0.86 & 0.69 & 0.61 & \textbf{98.12} \\
\bottomrule
\end{tabular}

%% file: tables/profile_ablation.tex
\begin{tabular}[t]{c|cc|cc|cc}
\toprule
\multirow{2}{*}{Method} & \multirow{2}{*}{$\Lambda$} & \multirow{2}{*}{$\mathcal{W}$} & \multicolumn{2}{c|}{\textbf{LLaVA-v1.5-13B}} & \multicolumn{2}{c}{\textbf{LLaVA-NeXT-13B}} \\
&  &  & FLOPs & Mem & FLOPs & Mem \\
\midrule
\rowcolor{gray!15} Vanilla & - & - & 19.4 & 0.55 & 65.0 & 1.75 \\
\midrule
\multirow{4}{*}{\rotatebox{90}{\ourmethod}} 
& \multirow{2}{*}{15,23,31} & 4,2,1 & 12.6 & 0.37 & 37.3 & 1.05 \\
&                           & 6,4,2 & 12.6 & 0.37 & 37.3 & 1.05 \\
& \multirow{2}{*}{17,24,31} & 4,2,1 & 13.1 & 0.38 & 39.0 & 1.09 \\
&                           & 6,4,2 & 13.1 & 0.38 & 39.0 & 1.09 \\



\bottomrule
\end{tabular}

%% file: sec/5_conclusion.tex
\section{Conclusion}
\label{sec:conclusion}

In this paper, we present \ourmethod, a novel \textit{training-free} approach for optimizing KV cache storage for general-purpose LVLMs.
It leverages \emph{text-prompt-guided graph message passing and aggregation} to informatively compress vision tokens during the \textit{prefill} stage of inference.
Our method is designed to be:
(i) memory-efficient: by progressively and dynamically compressing vision tokens through a hierarchical process; and
(ii) compute-efficient: by employing window-based graph partitioning and bipartite matching to accelerate message aggregation.
The experimental results demonstrate that our approach:
(a) largely preserves the general-purpose performance of the base LVLM across multiple benchmarks, and
(b) outperforms existing baselines in performance-efficiency trade-off.

\paragraph{Limitations} We acknowledge two limitations:
(a) \ourmethod\ leverages a bipartite graph matching algorithm, which splits vision tokens into two disjoint sets, then finds low-FD pairings between nodes across the two sets.
This limits the compression rate to a maximum of 50\% per step, thus requiring multiple iterations to achieve higher overall reduction.
(b) Furthermore, our method explicitly computes attention matrices for cross-modality guidance during a \emph{small number} of compression steps, similar to prior approaches~\citep{chenImageWorth12024, liuVisualInstructionTuning2023}.
These steps are less compatible with IO-efficient implementations such as FlashAttention~\citep{daoFlashAttentionFastMemoryEfficient2022}, which do not expose the full attention matrix.
However, layers where compression is not applied remain fully compatible with FlashAttention.

\section*{Acknowledgments}
We gratefully acknowledge the support of the NUS Artificial Intelligence Institute (NAII) through seed grant number NAII-SG-2025-027.

%% file: sec/X_suppl.tex
\clearpage
\appendix

\section{Appendix}

\subsection{Summary of notations}
Table~\ref{tab:appendix-notation-def} provides an overview of the notations used in this paper.
\begin{table}[ht]
    \centering
    \caption{Summary of notations.}
    \label{tab:appendix-notation-def}
    \begin{adjustbox}{max width=\linewidth}
        \input{tables/notation_def}
    \end{adjustbox}
\end{table}

\subsection{Method}

\subsubsection{Method overview}

\begin{figure}[!ht]
    \centering
    \includegraphics[width=0.5\linewidth]{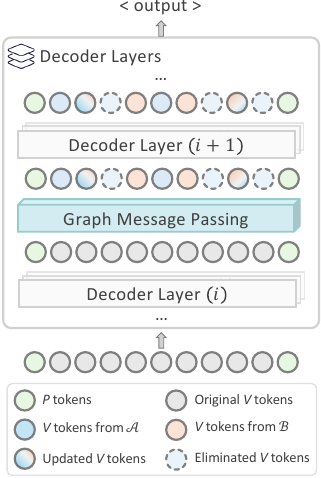}
    \caption{\ourmethod\ dynamically compresses vision tokens between two consecutive LVLM decoder layers. The key and value tokens are compressed simultaneously for later layers, reducing the memory used by KV cache.}
    \label{fig:supp-method-overview}
\end{figure}

As illustrated in Fig.~\ref{fig:supp-method-overview}, we insert graph message passing-based compression between two selected decoder layers in the LVLM, simultaneously reducing  KV cache size and the number of vision tokens processed by downstream layers.
Compression is performed three times to achieve the overall compression ratio.

\subsubsection{Adjacency matrix}
\label{sec:supp-adj-matrix}
In Sec.~\ref{sec:method-overview}, we defined for our bipartite graph with edges $M \in \{0,1\}^{|\mathcal{X}_\mathcal{A}|\times|\mathcal{X}_\mathcal{B}|}$, whose rows correspond to nodes in $\mathcal{X}_\mathcal{A}$ and columns to nodes in
$\mathcal{X}_\mathcal{B}$.
However, as the two subsets need not contain the same number of nodes,
$M$ is generally rectangular.  
Conventionally, for a standard graph, the adjacency matrix is square with side length equal to the total number of nodes.
The analogous square adjacency matrix for our bipartite graph is:
\begin{equation}
    \begin{pmatrix}
        0        & M \\
        M^{\top} & 0
    \end{pmatrix},
\end{equation}
where the upper-left and lower-right blocks are zero by definition.
Throughout our paper, we work directly with $M$, as this rectangular form is sufficient for message passing between the two partitions.

\subsection{Additional results}

\subsubsection{Additional backbones}

\paragraph{Qwen2.5-VL} We also evaluated \ourmethod\ on Qwen2.5-VL-7B-Instruct~\citep{baiQwen25VLTechnicalReport2025} across multiple compression ratios.
The results in Table~\ref{tab:main-results-qwen} demonstrate that \ourmethod\ yields substantial improvements compared to baseline approaches, preserving accuracy more effectively and delivering stronger overall performance under compression.
Notably, as presented in Table~\ref{tab:compress-results-qwen}, at more aggressive compression ratios, \ourmethod\ still delivers near-identical performance to the vanilla model.

\begin{table}[!th]
    \centering
    \caption{Results of \ourmethod\ on Qwen2.5-VL-7B-Instruct model at $55\%$ vision token retention in the KV cache. \textbf{Avg \%} denotes the average of all performance metrics normalized against the Vanilla model. Methods are then sorted by average score. ``NC'' and ``VW'' denote NoCaps and VizWiz, respectively.}
    \label{tab:main-results-qwen}
    \vspace{0.5em}
    \begin{adjustbox}{max width=\linewidth}
        \normalsize  
        \input{tables/qwen_55}
    \end{adjustbox}
\end{table}

\begin{table}[!th]
    \centering
    \caption{Results of \ourmethod\ on Qwen2.5-VL-7B-Instruct model at various retention rates of vision tokens in the KV cache. \textbf{Avg \%} denotes the average of all performance metrics normalized against the Vanilla model. ``NC'' and ``VW'' denote NoCaps and VizWiz, respectively.}
    \label{tab:compress-results-qwen}
    \vspace{0.5em}
    \begin{adjustbox}{max width=\linewidth}        
        \normalsize  
        \input{tables/qwen_compression_rates}
    \end{adjustbox}
\end{table}

\subsubsection{Additional ablation studies}
\label{sec:app-ablation}

\paragraph{Bipartite vs. full pairwise matching}
We provide additional ablation studies to analyze the design choice of bipartite matching compared to full pairwise matching. We evaluate both approaches from two perspectives: (1) downstream task performance and (2) computational efficiency.

While bipartite matching does not guarantee globally optimal pair assignments, we empirically observe that its impact on downstream performance is marginal.
The results are shown in Table~\ref{tab:ablation-bipartite_vs_pairwise}: across all benchmarks, bipartite matching achieves comparable performance to full pairwise matching.

We hypothesize that this behavior is due to our multi-stage compression strategy. Although globally optimal pairs may not be matched in early stages (\eg, when tokens fall into the same partition), these tokens are likely to be reassigned into different partitions in later stages, where they can then be matched and merged. This progressively mitigates the sub-optimality introduced by bipartite partitioning.

\begin{table}[t]
    \centering
    \caption{Performance comparison between bipartite matching and full pairwise matching on LLaVA-v1.5 when retaining 55\% of vision tokens.}
    \label{tab:ablation-bipartite_vs_pairwise}
    \begin{adjustbox}{width=0.6\linewidth}
        \input{tables/ablation_bipartite_vs_full}
    \end{adjustbox}
\end{table}

\begin{table}[t]
    \centering
    \caption{FLOPs comparison of bipartite and full pairwise matching across vision-token counts.}
    \label{tab:ablation-bipartite_vs_pairwise_flops}
    \begin{adjustbox}{max width=\linewidth}
        \input{tables/ablation_bipartite_vs_full_flops}
    \end{adjustbox}
\end{table}

We further compare the computational cost of the two matching strategies. As derived in Sec.~\ref{sec:intra-window-compression}, bipartite matching reduces the number of similarity comparisons from $\mathcal{O}(v_w^2/2)$ to $\mathcal{O}(v_w^2/4)$, effectively halving the pairwise operations.
In practice, however, we observe an even larger gap in runtime cost.
As shown in Table~\ref{tab:ablation-bipartite_vs_pairwise_flops}, full pairwise matching incurs approximately $4\times$ higher FLOPs than bipartite matching across different numbers of vision tokens. This is due to additional overhead in computing and maintaining the full similarity matrix. The increased computation also translates to higher memory (VRAM) usage.

Overall, bipartite matching provides a favorable trade-off between performance and efficiency.

\paragraph{Influence of window schedule}
Table~\ref{tab:ablation-window-next} studies the effect of window schedule $\mathcal{W}$, which is closely related to the number of vision tokens used by the LVLM.
A larger initial window size is appropriate when the model encodes images at high resolution, e.g., LLaVA-NeXT encodes an image into 2,144 tokens.
In contrast, a smaller value of $w$ is more favorable when there are fewer vision tokens, e.g., LLaVA-v1.5, which uses 576 vision tokens per image.
In our experiments, we used $\mathcal{W}=[6,4,2]$ for LLaVA-NeXT and $\mathcal{W}=[4,2,1]$ for LLaVA-v1.5.
We found that using a large window size with fewer vision tokens overly restricts token matching, often resulting in mismatches.
\begin{table}[!t]
    \centering
    \caption{Performance comparison across various combinations of $\mathcal{W}$ on LLaVA-13B models at $55\%$ vision token retention. ``NC'' and ``VW'' denote NoCaps and VizWiz, respectively.}
    \label{tab:ablation-window-next}
    \vspace{0.5em}
    \begin{adjustbox}{max size=\linewidth}
        \normalsize  
        \input{tables/ablation_window_next}
    \end{adjustbox}
\end{table}

\begin{figure}[!t]
    \centering
    \includegraphics[width=0.6\linewidth]{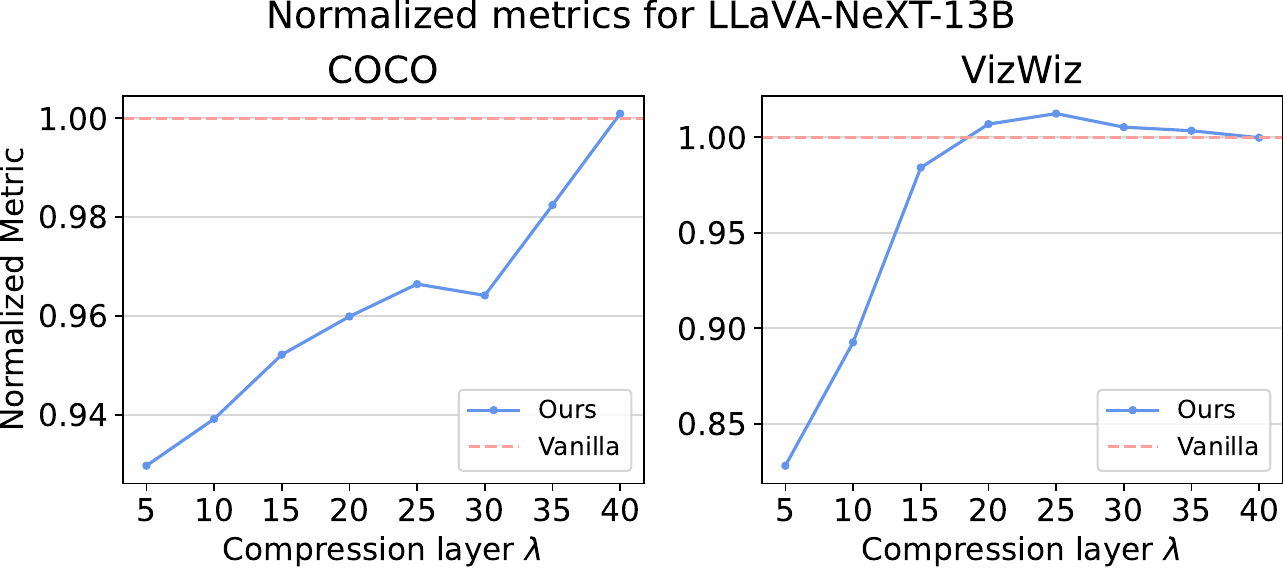}
    \caption{Performance comparison on LLaVA-NeXT-13B under different compression layer choices $\lambda$.}
    \label{fig:layer-ablation}
\end{figure}

\paragraph{Influence of compression layers}
We investigate the impact of varying layers for token compression, as illustrated in Fig.~\ref{fig:layer-ablation}.
Trends between the compression layer and model performance reveal that compressing in the shallow layers has a more substantial impact on performance.
This effect is particularly pronounced in VizWiz, where LVLMs must refrain from answering (\eg, when the ground truth is ``unanswerable'').
Compression in the deeper layers yields performance nearly identical to the base LVLM models, but offers little reduction in memory usage.

\paragraph{Overall robustness to compression schedule}
The compression schedule in our method is designed heuristically rather than learned from data.
This choice is intentional: our goal is to provide a training-free, plug-and-play solution that can be readily applied to arbitrary LVLMs without incurring additional training cost or requiring a learned policy for schedule selection.
To promote generalization and avoid task-specific bias, we determine the schedule parameters using a subset of benchmarks (COCO and MME), and then fix them across all remaining tasks.
This protocol reduces the risk of implicitly overfitting the schedule to any particular evaluation setting.
While learning an adaptive scheduler is an interesting direction for future work, our results suggest that such complexity may not be necessary for strong performance.
As shown in Table~\ref{tab:ablation-schedule}, performance remains stable across a range of $\Lambda$ and $\mathcal{W}$ configurations, indicating that the method is robust to the choice of schedule.

\begin{table}[!t]
    \centering
    \caption{Performance comparison across different schedules by varying $\Lambda$ and $\mathcal{W}$ at $55\%$ vision token retention. Our method is robust to the layer and window schedules.}
    \label{tab:ablation-schedule}
    \vspace{0.5em}
    \begin{adjustbox}{max width=\linewidth}
        \normalsize  
        \input{tables/ablation_schedule}
    \end{adjustbox}
\end{table}

\paragraph{Influence of similarity metrics}
We evaluate the impact of different similarity metrics used in the token-pairing process of \ourmethod, as first described in Sec.~\ref{sec:intra-window-compression}.
Table~\ref{tab:ablation-similarity-metric} compares the results of cosine similarity to Euclidean distance and L2-Squared distance.
Overall, cosine similarity consistently achieves the best and most stable performance across benchmarks.
In contrast, Euclidean and L2-Squared distances lead to noticeable degradation, particularly on tasks such as SeedBench and MME.
Based on these observations, we adopt cosine similarity as the default metric for token pairing in \ourmethod.

\begin{table}[!t]
    \centering
    \caption{Performance comparison of using cosine similarity, Euclidean distance and L2-Squared distance at $55\%$ vision token retention.}
    \label{tab:ablation-similarity-metric}
    \begin{adjustbox}{width=0.7\linewidth}
        \normalsize
        \input{tables/ablation_similarity_metric}
    \end{adjustbox}
\end{table}

\subsubsection{Additional latency profiles}
\label{sec:supp-latency}
We evaluate model responsiveness using two latency metrics: time-to-first-token (TTFT) and generation latency for 100 tokens.
As shown in Table~\ref{tab:latency-results-std}, TTFT highlights the overhead of the prefilling stage and directly reflects user-perceived responsiveness, while generation latency characterizes decoding efficiency.
Together, these results provide a comprehensive view of both initial response delay and sustained throughput.
\begin{table}[!t]
    \centering
    \caption{Latency comparison across LLaVA models. TTFT {=} Time to First Token. Gen latency {=} latency for generating 100 tokens. Lower is better.}
    \label{tab:latency-results-std}
    \vspace{0.5em}
    \normalsize  
    \begin{adjustbox}{width=0.7\linewidth}
        \input{tables/latency_llava}
    \end{adjustbox}
\end{table}

\subsubsection{Performance comparison to FastV}
We provide additional experiments to ensure a fair comparison with the FastV baseline.
In our main experiments, we followed the default FastV configuration as described in its original implementation, where pruning is performed at an early transformer layer (specifically, layer index $K=2$).
While this is a key design choice of FastV, it may not fully reflect its best achievable performance under different configurations.

To account for this, we conduct a more comprehensive evaluation by varying the pruning layer $K \in \{1,2,4,8\}$.
To ensure a controlled comparison, we adjust the retention ratio $R$ such that all variants maintain the same overall retention rate of vision tokens in the KV cache (55\%).

The results are summarized in Table~\ref{tab:fastv_k_sweep}. Across both LLaVA-v1.5-7B and LLaVA-NeXT-7B, LightKV consistently achieves competitive or superior performance compared to FastV under different choices of $K$.
Notably, while certain configurations of FastV (e.g., larger $K$) can partially recover performance, they still do not consistently surpass LightKV under the same compression budget.

\begin{table}[!t]
    \centering
    \caption{Performance comparison between \ourmethod\ and FastV under different pruning layers $K$ at $55\%$ vision token retention.}
    \label{tab:fastv_k_sweep}
    \begin{adjustbox}{max width=\linewidth}
        \input{tables/ablation_fastv_sweep}
    \end{adjustbox}
\end{table}

\subsubsection{Visualization}
\begin{figure}[!th]
    \centering
    \includegraphics[width=\linewidth]{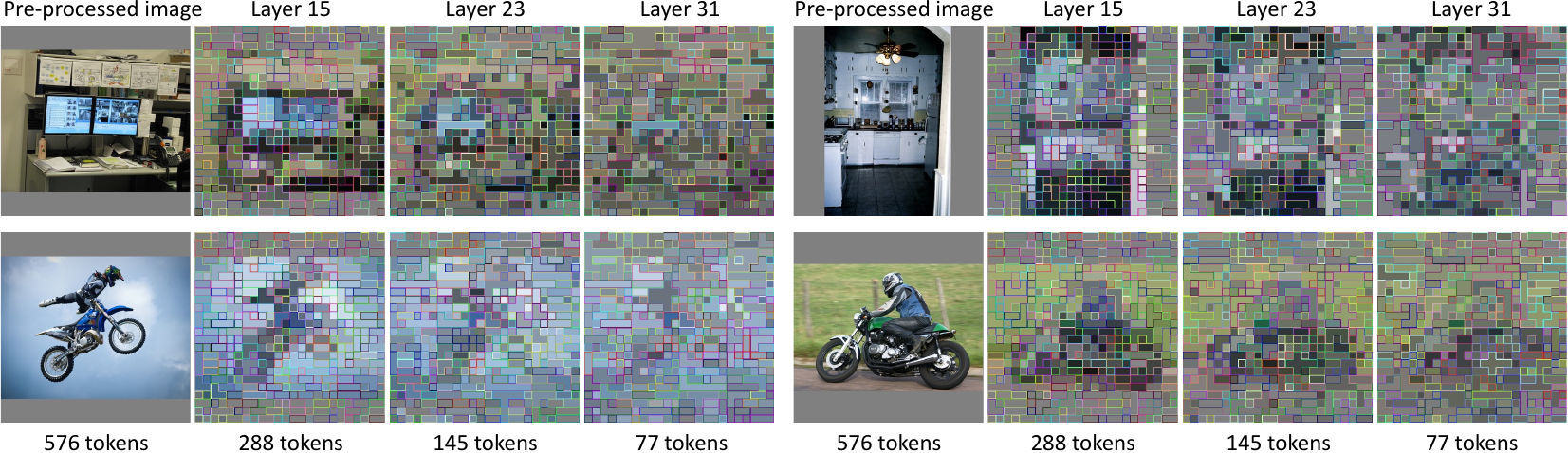}
    \caption{Visualization of a 3-stage vision token compression, halving tokens at each stage and achieving 55\% vision token retention in the KV cache. Distant patches may be compressed into a single token.}
    \label{fig:visualization}
    \vspace{-1em}
\end{figure}

We provide visualization cases for vision token compression of Coco images in Fig.~\ref{fig:visualization} for a 3-stage compression on LLaVA-v1.5-13B, reducing the number of tokens from $576 \rightarrow 288 \rightarrow 145 \rightarrow 77$.
Unlike conventional vision encoders, vision tokens in LVLMs incorporate prompt information.
As a result, visually similar patches may differ significantly in the embedding space, making it plausible to aggregate non-adjacent patches.
To this end, our intra-window strategy imposes constraints on this aggregation process to maintain spatial coherence during compression.

%% file: tables/notation_def.tex
\renewcommand{\arraystretch}{1.2}
\begin{tabular}{cl}
\toprule
\textbf{Notation} & \textbf{Definition} \\
\midrule
$L$                                     & Total number of decoder layers in the LVLM \\
$s$                                     & Number of compression iterations, with $s < L$ \\
$\lambda_i$                             & Decoder layer where the $i$-th compression step is performed \\
$\Lambda = [\lambda_1,\ldots,\lambda_s]$& Schedule of compression layers \\
$w_i$                                   & Number of window partitions \textit{along each axis} used at stage $i$ \\
$w_i^2$                                 & Number of window partitions used at stage $i$ \\
$\mathcal{W} = [w_1,\ldots,w_s]$        & Schedule of window partitions, with $w_i > w_{i+1}$ \\
$\rho_i$                                & Compression ratio applied at stage $i$ \\
$\mathcal{P} = [\rho_1,\ldots,\rho_s]$  & Schedule of compression rates \\
$X_v$                                   & Sequence / set of vision tokens \\
$v$                                     & Number of vision tokens \\
$\omega$                                & Index for an individual window \\
$v_\omega = \frac{v}{w \times w}$       & Number of vision tokens in window $\omega$ \\
$\mathbf{x}$                            & Embedding of a vision token / graph node \\
$\mathcal{X}=\{\mathbf{x} \mid \mathbf{x} \in X_v\}$      & Set of graph nodes formed from vision tokens \\
$\mathcal{X}_A, \mathcal{X}_B$          & Two near-equal subsets of nodes in the bipartite graph \\
$\mathcal{E} = \mathcal{X}_A \times \mathcal{X}_B$  & Edges of the bipartite graph \\
$M \in \{0,1\}^{|\mathcal{X}_A|\times |\mathcal{X}_B|}$ & Rectangular adjacency matrix for selected bipartite edges \\
$\mathrm{FD}(\alpha,\beta)$             & Feature divergence between two nodes with indices $\alpha$ and $\beta$ \\
$\mathcal{T}_\rho$                      & Set of the $\lfloor \rho v_{\omega}/2 \rfloor$ selected token pairs with smallest FD values \\
$\mathcal{X}_R$                         & Unconnected nodes left unchanged after edge selection \\
$A \in \mathbb{R}^{H \times (p+v)\times(p+v)}$
                                        & Multi-head attention matrix during prefill \\
$H$                                     & Number of attention heads \\
$p$                                     & Number of prompt tokens \\
$\mathcal{J}$                           & Index set of prompt tokens \\
$\xi_i$                                 & Prompt-guidance weight / accumulated attention for vision token $i$ \\
$\xi_A, \xi_B$                          & Prompt-guidance weights corresponding to $X_A$ and $X_B$ \\

\bottomrule
\end{tabular}

%% file: tables/qwen_55.tex
\begin{tabular}{l|ccccccccc|c}
\toprule
\multirow{2}{*}{Method} & 
  \multirow{2}{*}{Coco} & 
  \multirow{2}{*}{GQA} & 
  \multicolumn{2}{c}{MME} & 
  \multirow{2}{*}{NC} & 
  \multicolumn{2}{c}{Pope} & 
  \multirow{2}{*}{Seed} & 
  \multirow{2}{*}{VW} & 
  \multirow{2}{*}{\textbf{Avg \%}} \\
\cmidrule(l){4-5} \cmidrule(l){7-8}
& & & \multicolumn{1}{c}{C} & \multicolumn{1}{c}{P} & & \multicolumn{1}{c}{Acc} & \multicolumn{1}{c}{F1} & & & \\
  
\midrule
\rowcolor{gray!15} Vanilla & 0.319 & 0.604 & 638.21 & 1695.25 & 0.372 & 0.875 & 0.862 & 0.790 & 0.704 & 100.00 \\
FastV      & 0.339 & 0.587 & 625.35 & 1687.78 & 0.386 & 0.869 & 0.853 & 0.744 & 0.698 & 98.77 \\
ToMe       & 0.329 & 0.591 & 640.71 & 1687.75 & 0.425 & 0.862 & 0.782 & 0.782 & 0.683 & 100.04 \\
PiToMe     & 0.389 & 0.584 & 624.64 & 1671.09 & 0.433 & 0.860 & 0.842 & 0.774 & 0.691 & 100.24 \\
ToFu       & 0.383 & 0.587 & 657.86 & 1683.05 & 0.418 & 0.857 & 0.839 & 0.788 & 0.696 & 100.75 \\
\rowcolor{cyan!10} \ourmethod & 0.389 & 0.591 & 647.50 & 1706.38 & 0.435 & 0.863 & 0.846 & 0.780 & 0.694 & \textbf{101.37} \\

\bottomrule
\end{tabular}

%% file: tables/qwen_compression_rates.tex
\begin{tabular}{l|ccccccccc|c}
\toprule
\multirow{2}{*}{Rate} &
  \multirow{2}{*}{Coco} & 
  \multirow{2}{*}{GQA} & 
  \multicolumn{2}{c}{MME} & 
  \multirow{2}{*}{NC} & 
  \multicolumn{2}{c}{Pope} & 
  \multirow{2}{*}{Seed} & 
  \multirow{2}{*}{VW} & 
  \multirow{2}{*}{\textbf{Avg \%}} \\
\cmidrule(l){4-5} \cmidrule(l){7-8}
& & & \multicolumn{1}{c}{C} & \multicolumn{1}{c}{P} & & \multicolumn{1}{c}{Acc} & \multicolumn{1}{c}{F1} & & & \\
\midrule
\rowcolor{gray!15} Vanilla & 0.319 & 0.604 & 638.21 & 1695.25 & 0.372 & 0.875 & 0.862 & 0.790 & 0.704 & 100.00 \\

55\% & 0.389 & 0.591 & 647.50 & 1706.38 & 0.435 & 0.863 & 0.846 & 0.780 & 0.694 & 101.37 \\

40\% & 0.370 & 0.586 & 611.78 & 1632.64 & 0.450 & 0.851 & 0.830 & 0.754 & 0.666 & 101.01 \\

30\% & 0.361 & 0.581 & 588.93 & 1574.34 & 0.455 & 0.833 & 0.806 & 0.732 & 0.670 & 98.89 \\

20\% & 0.356 & 0.569 & 591.78 & 1612.83 & 0.458 & 0.835 & 0.809 & 0.730 & 0.667 & 99.24 \\

\bottomrule
\end{tabular}

%% file: tables/ablation_bipartite_vs_full.tex
\begin{tabular}{l|cccc|c}
\toprule
\multirow{2}{*}{Method} 
& \multicolumn{2}{c|}{MME} 
& \multicolumn{2}{c|}{Pope} 
& \multirow{2}{*}{Avg\ \%} \\
\cmidrule(lr){2-3} \cmidrule(lr){4-5}
& C & P & Acc. & F1 & \\
\midrule
\multicolumn{6}{c}{\textbf{LLaVA-v1.5-13B}} \\
Bipartite       & 302.1	& 1543.8 & 0.87 & 0.86 & 100.75 \\
Full Pairwise   & 298.9	& 1532.0 & 0.87 & 0.86 & 100.31 \\
\midrule
\multicolumn{6}{c}{\textbf{LLaVA-v1.5-7B}} \\
Bipartite       & 357.5 & 1519.8 & 0.87 & 0.86 & 100.30 \\
Full Pairwise   & 371.0 & 1522.2 & 0.86 & 0.85 & 100.71 \\
\bottomrule
\end{tabular}

%% file: tables/ablation_bipartite_vs_full_flops.tex
\begin{tabular}{l|cccc}
\toprule
\multicolumn{1}{l|}{\multirow{2}{*}{Method}} & \multicolumn{4}{c}{\# Vision tokens} \\ 
\cmidrule(lr){2-5}
\multicolumn{1}{l|}{} & 512 & 1024 & 2048 & 4096 \\
\midrule
Bipartite     & 0.134 & 0.538 & 2.151 & 8.602         \\
Full Pairwise & 0.538 & 2.151 & 8.603 & 34.410        \\
\bottomrule
\end{tabular}

%% file: tables/ablation_window_next.tex
\begin{tabular}{l|ccccccccc|c}
\toprule
\multirow{2}{*}{Method} & 
  \multirow{2}{*}{Coco} & 
  \multicolumn{2}{c}{MME} & 
  \multirow{2}{*}{NC} & 
  \multicolumn{2}{c}{Pope} & 
  \multirow{2}{*}{SQA} & 
  \multirow{2}{*}{Seed} & 
  \multirow{2}{*}{VW} & 
  \multirow{2}{*}{\textbf{Avg \%}} \\ 
\cmidrule(l){3-4} \cmidrule(l){6-7}
& & \multicolumn{1}{c}{C} & \multicolumn{1}{c}{P} & & \multicolumn{1}{c}{Acc} & \multicolumn{1}{c}{F1} & & & \\
\midrule
\multicolumn{11}{c}{\textbf{LLaVA-v1.5-13B}} \\
\rowcolor{gray!15} Vanilla & 1.16 & 295.36 & 1532.0 & 1.09 & 0.87 & 0.86 & 0.73 & 0.69 & 0.57 & 100.00 \\
LightKV $\mathcal{W}{=}$[4,2,1]   & 1.15 & 302.14 & 1543.8 & 1.08 & 0.87 & 0.86 & 0.72 & 0.69 & 0.56 & 99.80 \\
LightKV $\mathcal{W}{=}$[6,4,2]   & 1.14 & 301.79 & 1541.1 & 1.08 & 0.87 & 0.86 & 0.72 & 0.69 & 0.56 & 99.67 \\
\midrule
\multicolumn{11}{c}{\textbf{LLaVA-NeXT-13B}} \\
\rowcolor{gray!15} Vanilla & 1.02 & 318.93 & 1575.1 & 0.88 & 0.88 & 0.86 & 0.73 & 0.69 & 0.64 & 100.00 \\
LightKV $\mathcal{W}{=}$[4,2,1]   & 0.96 & 311.43 & 1576.3 & 0.83 & 0.87 & 0.86 & 0.59 & 0.69 & 0.61 & 95.68 \\
LightKV $\mathcal{W}{=}$[6,4,2]   & 0.96 & 326.07 & 1576.5 & 0.83 & 0.87 & 0.86 & 0.59 & 0.69 & 0.61 & 96.20 \\
\bottomrule
\end{tabular}

%% file: tables/ablation_schedule.tex
\begin{tabular}{cc|cccccccc|c}
\toprule
\multirow{2}{*}{$\Lambda$} & \multirow{2}{*}{$\mathcal{W}$} &
\multirow{2}{*}{Coco} &
\multicolumn{2}{c}{MME} &
\multirow{2}{*}{NC} &
\multicolumn{2}{c}{Pope} &
\multirow{2}{*}{Seed} &
\multirow{2}{*}{VW} &
\multirow{2}{*}{\textbf{Avg \%}} \\
\cmidrule(l){4-5} \cmidrule(l){7-8}
& & & C & P & & Acc & F1 & & & \\
\midrule
\multicolumn{11}{c}{\textbf{LLaVA-v1.5-13B}} \\
$15,23,31$ & $4,2,1$ & 1.15 & 302.1 & 1543.8 & 1.08 & 0.87 & 0.86 & 0.69 & 0.56 & \textbf{99.94} \\
$15,23,31$ & $6,4,2$ & 1.15 & 295.5 & 1530.8 & 1.08 & 0.88 & 0.86 & 0.69 & 0.57 & 99.92 \\
$9,19,29$ & $4,2,1$  & 1.17 & 263.6 & 1518.9 & 1.07 & 0.87 & 0.86 & 0.69 & 0.52 & 97.32 \\
$9,19,29$ & $6,4,2$  & 1.17 & 263.7 & 1518.9 & 1.07 & 0.87 & 0.86 & 0.68 & 0.53 & 97.38 \\
\midrule
\multicolumn{11}{c}{\textbf{LLaVA-v1.5-7B}} \\
$12,18,24$ & $6,4,2$ & 1.11 & 357.5 & 1519.8 & 1.03 & 0.87 & 0.86 & 0.66 & 0.53 & \textbf{99.79} \\
$12,18,24$ & $4,2,1$ & 1.11 & 354.3 & 1512.6 & 1.04 & 0.87 & 0.85 & 0.67 & 0.53 & 99.66 \\
\bottomrule
\end{tabular}


%% file: tables/ablation_similarity_metric.tex
\begin{tabular}{l|cccccccc|c}
\toprule
\multirow{2}{*}{Metric} &
\multirow{2}{*}{Coco} &
\multicolumn{2}{c}{MME} &
\multirow{2}{*}{NC} &
\multicolumn{2}{c}{Pope} &
\multirow{2}{*}{Seed} &
\multirow{2}{*}{VW} &
\multirow{2}{*}{\textbf{Avg \%}} \\
\cmidrule(l){3-4} \cmidrule(l){6-7}
& & C & P & & Acc & F1 & & & \\
\midrule
\multicolumn{10}{c}{\textbf{LLaVA-v1.5-13B}} \\
Euclidean & 1.15 & 297.5 & 1529.2 & 1.07 & 0.87 & 0.86 & 0.39 & 0.55 & 93.85 \\
L2-Squared & 1.15 & 292.5 & 1530.0 & 1.08 & 0.87 & 0.86 & 0.39 & 0.55 & 93.77 \\
\rowcolor{cyan!10} Cosine & 1.15 & 302.1 & 1543.8 & 1.08 & 0.87 & 0.86 & 0.69 & 0.56 & \textbf{99.94} \\
\midrule
\multicolumn{10}{c}{\textbf{LLaVA-NeXT-13B}} \\
Euclidean & 0.96 & 316.0 & 1553.9 & 0.84 & 0.87 & 0.86 & 0.34 & 0.59 & 90.96 \\
L2-Squared & 0.98 & 308.5 & 1554.7 & 0.84 & 0.87 & 0.86 & 0.35 & 0.60 & 91.29 \\
\rowcolor{cyan!10} Cosine & 0.96 & 326.1 & 1576.5 & 0.83 & 0.87 & 0.86 & 0.69 & 0.61 & \textbf{98.12} \\
\bottomrule
\end{tabular}

%% file: tables/latency_llava.tex
\begin{tabular}{l|cc|cc}
\toprule
Method & TTFT (ms) & Gen latency (s) & TTFT (ms) & Gen latency (s) \\
\midrule
& \multicolumn{2}{c}{\textbf{LLaVA-v1.5-13B}} & \multicolumn{2}{c}{\textbf{LLaVA-v1.5-7B}} \\
\rowcolor{gray!15} Vanilla  & 130 $\pm$ 0.393 & 2.89 $\pm$ 0.004 & 78 $\pm$ 0.827 & 2.09 $\pm$ 0.008 \\
FastV    &  85 $\pm$ 0.450 & 2.59 $\pm$ 0.003 & 52 $\pm$ 0.316 & 1.85 $\pm$ 0.004 \\
PiToMe   &  93 $\pm$ 0.768 & 2.79 $\pm$ 0.017 & 58 $\pm$ 1.543 & 2.14 $\pm$ 0.010 \\
ToFu     &  94 $\pm$ 1.152 & 2.83 $\pm$ 0.009 & 58 $\pm$ 0.428 & 2.14 $\pm$ 0.004 \\
ToMe (P) &  94 $\pm$ 0.278 & 2.77 $\pm$ 0.003 & 58 $\pm$ 0.647 & 2.13 $\pm$ 0.003 \\
\rowcolor{cyan!10} LightKV  &  98 $\pm$ 1.045 & 2.80 $\pm$ 0.004 & 65 $\pm$ 3.054 & 2.11 $\pm$ 0.006 \\
\midrule
& \multicolumn{2}{c}{\textbf{LLaVA-NeXT-13B}} & \multicolumn{2}{c}{\textbf{LLaVA-NeXT-7B}} \\
\rowcolor{gray!15} Vanilla  & 656 $\pm$ 1.479 & 3.79 $\pm$ 0.013 & 397 $\pm$ 1.337 & 2.43 $\pm$ 0.005 \\
FastV    & 321 $\pm$ 2.142 & 3.03 $\pm$ 0.003 & 197 $\pm$ 1.207 & 2.03 $\pm$ 0.004 \\
PiToMe   & 396 $\pm$ 1.109 & 3.30 $\pm$ 0.004 & 247 $\pm$ 0.633 & 2.34 $\pm$ 0.005 \\
ToFu     & 394 $\pm$ 0.968 & 3.30 $\pm$ 0.003 & 245 $\pm$ 0.757 & 2.35 $\pm$ 0.005 \\
ToMe (P) & 394 $\pm$ 1.280 & 3.30 $\pm$ 0.004 & 245 $\pm$ 0.722 & 2.32 $\pm$ 0.002 \\
\rowcolor{cyan!10} LightKV  & 383 $\pm$ 0.878 & 3.29 $\pm$ 0.002 & 259 $\pm$ 0.643 & 2.30 $\pm$ 0.004 \\
\bottomrule
\end{tabular}


%% file: tables/ablation_fastv_sweep.tex
\begin{tabular}{cl|cccccccc|r}
\toprule
\multicolumn{2}{c|}{\multirow{2}{*}{Method}} & 
  \multirow{2}{*}{Coco} & 
  \multicolumn{2}{c}{MME} & 
  \multirow{2}{*}{NC} & 
  \multicolumn{2}{c}{Pope} & 
  \multirow{2}{*}{Seed} & 
  \multirow{2}{*}{VW} & 
  \multirow{2}{*}{\textbf{Avg \%}} \\ 
\cmidrule(l){4-5} \cmidrule(l){7-8}
& & & \multicolumn{1}{c}{C} & \multicolumn{1}{c}{P} & & \multicolumn{1}{c}{Acc} & \multicolumn{1}{c}{F1} & & & \\
\midrule

\multicolumn{11}{c}{\textbf{LLaVA-v1.5-7B}} \\
\multirow{4}{*}{\rotatebox{90}{FastV}} 
& $K{=}1$ & 1.08 & 337.8 & 1469.7 & 1.02 & 0.83 & 0.81 & 0.57 & 0.54 & 95.45 \\
& $K{=}2$ & 1.10 & 351.1 & 1513.7 & 1.04 & 0.85 & 0.83 & 0.66 & 0.54 & 99.03 \\
& $K{=}4$ & 1.10 & 339.2 & 1500.0 & 1.04 & 0.84 & 0.81 & 0.66 & 0.54 & 98.06 \\
& $K{=}8$ & 1.10 & 371.7 & 1502.0 & 1.02 & 0.85 & 0.84 & 0.65 & 0.53 & \underline{99.13} \\
\rowcolor{cyan!10} 
\multicolumn{2}{c|}{\ourmethod} & 1.11 & 357.5 & 1519.8 & 1.03 & 0.87 & 0.86 & 0.66 & 0.53 & \textbf{99.79} \\

\midrule
\multicolumn{11}{c}{\textbf{LLaVA-NeXT-7B}} \\
\multirow{4}{*}{\rotatebox{90}{FastV}} 
& $K{=}1$ & 0.96 & 326.0 & 1495.2 & 0.85 & 0.86 & 0.84 & 0.68 & 0.60 & \underline{97.87} \\
& $K{=}2$ & 0.88 & 265.4 & 1341.3 & 0.78 & 0.81 & 0.77 & 0.69 & 0.58 & 90.37 \\
& $K{=}4$ & 0.98 & 298.9 & 1504.5 & 0.86 & 0.87 & 0.85 & 0.68 & 0.59 & 97.38 \\
& $K{=}8$ & 0.96 & 293.2 & 1505.7 & 0.84 & 0.87 & 0.84 & 0.67 & 0.60 & 96.52 \\
\rowcolor{cyan!10} 
\multicolumn{2}{c|}{\ourmethod} & 0.98 & 338.6 & 1517.3 & 0.83 & 0.88 & 0.86 & 0.69 & 0.58 & \textbf{98.85} \\
\bottomrule
\end{tabular}

%% file: main.bib
@inproceedings{agrawalNocapsNovelObject2019,
  title = {Nocaps: Novel Object Captioning at Scale},
  shorttitle = {Nocaps},
  booktitle = {CVPR},
  author = {Agrawal, Harsh and Desai, Karan and Wang, Yufei and Chen, Xinlei and Jain, Rishabh and Johnson, Mark and Batra, Dhruv and Parikh, Devi and Lee, Stefan and Anderson, Peter},
  year = 2019,
  pages = {8948--8957}
}

@article{ainslieGQATrainingGeneralized2023,
  title = {GQA: Training Generalized Multi-Query Transformer Models from Multi-Head Checkpoints},
  shorttitle = {GQA},
  author = {Ainslie, Joshua and {Lee-Thorp}, James and de Jong, Michiel and Zemlyanskiy, Yury and Lebr{\'o}n, Federico and Sanghai, Sumit},
  year = 2023,
  primaryclass = {cs},
  doi = {10.48550/arXiv.2305.13245},
  archiveprefix = {arXiv},
  journal = {arXiv preprint arXiv:2305.13245}
}

@inproceedings{alayracFlamingoVisualLanguage2022,
  title = {Flamingo: A Visual Language Model for Few-Shot Learning},
  shorttitle = {Flamingo},
  booktitle = {NeurIPS},
  author = {Alayrac, Jean-Baptiste and Donahue, Jeff and Luc, Pauline and Miech, Antoine and Barr, Iain and Hasson, Yana and Lenc, Karel and Mensch, Arthur and Millican, Katherine and Reynolds, Malcolm and Ring, Roman and Rutherford, Eliza and Cabi, Serkan and Han, Tengda and Gong, Zhitao and Samangooei, Sina and Monteiro, Marianne and Menick, Jacob L. and Borgeaud, Sebastian and Brock, Andy and Nematzadeh, Aida and Sharifzadeh, Sahand and Bi{\'n}kowski, Miko{\l}aj and Barreira, Ricardo and Vinyals, Oriol and Zisserman, Andrew and Simonyan, Kar{\'e}n},
  year = 2022,
  pages = {23716--23736}
}

@inproceedings{alvarDivPruneDiversitybasedVisual2025,
  title = {DivPrune: Diversity-Based Visual Token Pruning for Large Multimodal Models},
  shorttitle = {DivPrune},
  booktitle = {CVPR},
  author = {Alvar, Saeed Ranjbar and Singh, Gursimran and Akbari, Mohammad and Zhang, Yong},
  year = 2025,
  eprint = {2503.02175},
  primaryclass = {cs},
  doi = {10.48550/arXiv.2503.02175},
  archiveprefix = {arXiv}
}

@inproceedings{arifHiREDAttentionguidedToken2025,
  title = {HiRED: Attention-Guided Token Dropping for Efficient Inference of High-Resolution Vision-Language Models},
  shorttitle = {HiRED},
  booktitle = {AAAI},
  author = {Arif, Kazi Hasan Ibn and Yoon, JinYi and Nikolopoulos, Dimitrios S. and Vandierendonck, Hans and John, Deepu and Ji, Bo},
  year = 2025,
  series = {AAAI'25/IAAI'25/EAAI'25},
  volume = {39},
  pages = {1773--1781},
  doi = {10.1609/aaai.v39i2.32171},
  isbn = {978-1-57735-897-8}
}

@article{awadallaOpenFlamingoOpenSourceFramework2023,
  title = {OpenFlamingo: An Open-Source Framework for Training Large Autoregressive Vision-Language Models},
  shorttitle = {OpenFlamingo},
  author = {Awadalla, Anas and Gao, Irena and Gardner, Josh and Hessel, Jack and Hanafy, Yusuf and Zhu, Wanrong and Marathe, Kalyani and Bitton, Yonatan and Gadre, Samir and Sagawa, Shiori and Jitsev, Jenia and Kornblith, Simon and Koh, Pang Wei and Ilharco, Gabriel and Wortsman, Mitchell and Schmidt, Ludwig},
  year = 2023,
  primaryclass = {cs},
  doi = {10.48550/arXiv.2308.01390},
  archiveprefix = {arXiv},
  journal = {arXiv preprint arXiv:2308.01390}
}

@article{baiQwen25VLTechnicalReport2025,
  title = {Qwen2.5-VL Technical Report},
  author = {Bai, Shuai and Chen, Keqin and Liu, Xuejing and Wang, Jialin and Ge, Wenbin and Song, Sibo and Dang, Kai and Wang, Peng and Wang, Shijie and Tang, Jun and Zhong, Humen and Zhu, Yuanzhi and Yang, Mingkun and Li, Zhaohai and Wan, Jianqiang and Wang, Pengfei and Ding, Wei and Fu, Zheren and Xu, Yiheng and Ye, Jiabo and Zhang, Xi and Xie, Tianbao and Cheng, Zesen and Zhang, Hang and Yang, Zhibo and Xu, Haiyang and Lin, Junyang},
  year = 2025,
  primaryclass = {cs},
  doi = {10.48550/arXiv.2502.13923},
  archiveprefix = {arXiv},
  journal = {arXiv preprint arXiv:2502.13923}
}

@article{baiQwenVLVersatileVisionLanguage2023,
  title = {Qwen-VL: A Versatile Vision-Language Model for Understanding, Localization, Text Reading, and Beyond},
  shorttitle = {Qwen-VL},
  author = {Bai, Jinze and Bai, Shuai and Yang, Shusheng and Wang, Shijie and Tan, Sinan and Wang, Peng and Lin, Junyang and Zhou, Chang and Zhou, Jingren},
  year = 2023,
  primaryclass = {cs},
  archiveprefix = {arXiv},
  journal = {arXiv preprint arXiv:2308.12966}
}

@inproceedings{baoBEiTBERTPreTraining2022,
  title = {BEiT: BERT Pre-Training of Image Transformers},
  shorttitle = {BEiT},
  booktitle = {ICLR},
  author = {Bao, Hangbo and Dong, Li and Piao, Songhao and Wei, Furu},
  year = 2022
}

@inproceedings{bolyaTokenMergingYour2023,
  title = {Token Merging: Your ViT But Faster},
  shorttitle = {ToMe},
  booktitle = {ICLR},
  author = {Bolya, Daniel and Fu, Cheng-Yang and Dai, Xiaoliang and Zhang, Peizhao and Feichtenhofer, Christoph and Hoffman, Judy},
  year = 2023
}

@article{bonnaerensLearnedThresholdsToken2023,
  title = {Learned Thresholds Token Merging and Pruning for Vision Transformers},
  shorttitle = {LTMP},
  author = {Bonnaerens, Maxim and Dambre, Joni},
  year = 2023,
  journal = {TMLR}
}

@article{caiPyramidKVDynamicKV2024,
  title = {PyramidKV: Dynamic KV Cache Compression Based on Pyramidal Information Funneling},
  shorttitle = {PyramidKV},
  author = {Cai, Zefan and Zhang, Yichi and Gao, Bofei and Liu, Yuliang and Liu, Tianyu and Lu, Keming and Xiong, Wayne and Dong, Yue and Chang, Baobao and Hu, Junjie and Xiao, Wen},
  year = 2024,
  archiveprefix = {arXiv},
  journal = {arXiv preprint arXiv:2406.02069}
}

@inproceedings{chenDiffRateDifferentiableCompression2023,
  title = {DiffRate : Differentiable Compression Rate for Efficient Vision Transformers},
  shorttitle = {DiffRate},
  booktitle = {ICCV},
  author = {Chen, Mengzhao and Shao, Wenqi and Xu, Peng and Lin, Mingbao and Zhang, Kaipeng and Chao, Fei and Ji, Rongrong and Qiao, Yu and Luo, Ping},
  year = 2023,
  pages = {17118--17128},
  issn = {2380-7504},
  doi = {10.1109/ICCV51070.2023.01574}
}

@inproceedings{chenEfficientLargeMultimodal2024,
  title = {Efficient Large Multi-Modal Models via Visual Context Compression},
  shorttitle = {LLaVolta},
  booktitle = {NeurIPS},
  author = {Chen, Jieneng and Ye, Luoxin and He, Ju and Wang, Zhao-Yang and Khashabi, Daniel and Yuille, Alan},
  year = 2024,
  pages = {73986--74007}
}

@article{chenExpandingPerformanceBoundaries2025,
  title = {Expanding Performance Boundaries of Open-Source Multimodal Models with Model, Data, and Test-Time Scaling},
  author = {Chen, Zhe and Wang, Weiyun and Cao, Yue and Liu, Yangzhou and Gao, Zhangwei and Cui, Erfei and Zhu, Jinguo and Ye, Shenglong and Tian, Hao and Liu, Zhaoyang and Gu, Lixin and Wang, Xuehui and Li, Qingyun and Ren, Yimin and Chen, Zixuan and Luo, Jiapeng and Wang, Jiahao and Jiang, Tan and Wang, Bo and He, Conghui and Shi, Botian and Zhang, Xingcheng and Lv, Han and Wang, Yi and Shao, Wenqi and Chu, Pei and Tu, Zhongying and He, Tong and Wu, Zhiyong and Deng, Huipeng and Ge, Jiaye and Chen, Kai and Zhang, Kaipeng and Wang, Limin and Dou, Min and Lu, Lewei and Zhu, Xizhou and Lu, Tong and Lin, Dahua and Qiao, Yu and Dai, Jifeng and Wang, Wenhai},
  year = 2025,
  primaryclass = {cs},
  doi = {10.48550/arXiv.2412.05271},
  archiveprefix = {arXiv},
  journal = {arXiv preprint arXiv:2412.05271}
}

@article{chenHowFarAre2024,
  title = {How Far Are We to GPT-4V? Closing the Gap to Commercial Multimodal Models with Open-Source Suites},
  shorttitle = {How Far Are We to GPT-4V?},
  author = {Chen, Zhe and Wang, Weiyun and Tian, Hao and Ye, Shenglong and Gao, Zhangwei and Cui, Erfei and Tong, Wenwen and Hu, Kongzhi and Luo, Jiapeng and Ma, Zheng and Ma, Ji and Wang, Jiaqi and Dong, Xiaoyi and Yan, Hang and Guo, Hewei and He, Conghui and Shi, Botian and Jin, Zhenjiang and Xu, Chao and Wang, Bin and Wei, Xingjian and Li, Wei and Zhang, Wenjian and Zhang, Bo and Cai, Pinlong and Wen, Licheng and Yan, Xiangchao and Dou, Min and Lu, Lewei and Zhu, Xizhou and Lu, Tong and Lin, Dahua and Qiao, Yu and Dai, Jifeng and Wang, Wenhai},
  year = 2024,
  primaryclass = {cs},
  doi = {10.48550/arXiv.2404.16821},
  archiveprefix = {arXiv},
  journal = {arXiv preprint arXiv:2404.16821}
}

@inproceedings{chenImageWorth12024,
  title = {An Image Is Worth 1/2 Tokens After Layer 2: Plug-and-Play Inference Acceleration for~Large Vision-Language Models},
  shorttitle = {An Image Is Worth 1/2 Tokens After Layer 2},
  booktitle = {ECCV},
  author = {Chen, Liang and Zhao, Haozhe and Liu, Tianyu and Bai, Shuai and Lin, Junyang and Zhou, Chang and Chang, Baobao},
  editor = {Leonardis, Ale{\v s} and Ricci, Elisa and Roth, Stefan and Russakovsky, Olga and Sattler, Torsten and Varol, G{\"u}l},
  year = 2024,
  pages = {19--35},
  doi = {10.1007/978-3-031-73004-7_2},
  isbn = {978-3-031-73004-7}
}

@inproceedings{chenInternVLScaling2024,
  title = {Intern VL: Scaling up Vision Foundation Models and Aligning for Generic Visual-Linguistic Tasks},
  shorttitle = {Intern VL},
  booktitle = {CVPR},
  author = {Chen, Zhe and Wu, Jiannan and Wang, Wenhai and Su, Weijie and Chen, Guo and Xing, Sen and Zhong, Muyan and Zhang, Qinglong and Zhu, Xizhou and Lu, Lewei and Li, Bin and Luo, Ping and Lu, Tong and Qiao, Yu and Dai, Jifeng},
  year = 2024,
  pages = {24185--24198},
  issn = {2575-7075},
  doi = {10.1109/CVPR52733.2024.02283}
}

@inproceedings{daiInstructBLIPGeneralpurposeVisionLanguage2023,
  title = {InstructBLIP: Towards General-Purpose Vision-Language Models with Instruction Tuning},
  shorttitle = {InstructBLIP},
  booktitle = {NeurIPS},
  author = {Dai, Wenliang and Li, Junnan and Li, Dongxu and Tiong, Anthony and Zhao, Junqi and Wang, Weisheng and Li, Boyang and Fung, Pascale N. and Hoi, Steven},
  year = 2023,
  pages = {49250--49267}
}

@inproceedings{daoFlashAttentionFastMemoryEfficient2022,
  title = {FlashAttention: Fast and Memory-Efficient Exact Attention with IO-Awareness},
  shorttitle = {FlashAttention},
  booktitle = {NeurIPS},
  author = {Dao, Tri and Fu, Dan and Ermon, Stefano and Rudra, Atri and R{\'e}, Christopher},
  year = 2022,
  pages = {16344--16359}
}

@inproceedings{diaoUnveilingEncoderfreeVisionlanguage2025,
  title = {Unveiling Encoder-Free Vision-Language Models},
  booktitle = {NeurIPS},
  author = {Diao, Haiwen and Cui, Yufeng and Li, Xiaotong and Wang, Yueze and Lu, Huchuan and Wang, Xinlong},
  year = 2025,
  pages = {52545--52567},
  isbn = {979-8-3313-1438-5}
}

@inproceedings{dosovitskiyImageWorth16x162021,
  title = {An Image Is Worth 16x16 Words: Transformers for Image Recognition at Scale},
  shorttitle = {ViT},
  booktitle = {ICLR},
  author = {Dosovitskiy, Alexey and Beyer, Lucas and Kolesnikov, Alexander and Weissenborn, Dirk and Zhai, Xiaohua and Unterthiner, Thomas and Dehghani, Mostafa and Minderer, Matthias and Heigold, Georg and Gelly, Sylvain and Uszkoreit, Jakob and Houlsby, Neil},
  year = 2021
}

@article{driessPaLMEEmbodiedMultimodal2023,
  title = {PaLM-E: An Embodied Multimodal Language Model},
  shorttitle = {PaLM-E},
  author = {Driess, Danny and Xia, Fei and Sajjadi, Mehdi S. M. and Lynch, Corey and Chowdhery, Aakanksha and Ichter, Brian and Wahid, Ayzaan and Tompson, Jonathan and Vuong, Quan and Yu, Tianhe and Huang, Wenlong and Chebotar, Yevgen and Sermanet, Pierre and Duckworth, Daniel and Levine, Sergey and Vanhoucke, Vincent and Hausman, Karol and Toussaint, Marc and Greff, Klaus and Zeng, Andy and Mordatch, Igor and Florence, Pete},
  year = 2023,
  primaryclass = {cs},
  archiveprefix = {arXiv},
  journal = {arXiv preprint arXiv:2303.03378}
}

@article{duHowGPTLearns2025,
  title = {How GPT Learns Layer by Layer},
  author = {Du, Jason and Hong, Kelly and Imran, Alishba and Jahanparast, Erfan and Khfifi, Mehdi and Qiao, Kaichun},
  year = 2025,
  primaryclass = {cs},
  doi = {10.48550/arXiv.2501.07108},
  archiveprefix = {arXiv},
  journal = {arXiv preprint arXiv:2501.07108}
}

@inproceedings{fayyazAdaptiveTokenSampling2022,
  title = {Adaptive Token Sampling for~Efficient Vision Transformers},
  shorttitle = {ATS},
  booktitle = {ECCV},
  author = {Fayyaz, Mohsen and Koohpayegani, Soroush Abbasi and Jafari, Farnoush Rezaei and Sengupta, Sunando and Joze, Hamid Reza Vaezi and Sommerlade, Eric and Pirsiavash, Hamed and Gall, J{\"u}rgen},
  year = 2022,
  pages = {396--414},
  doi = {10.1007/978-3-031-20083-0_24},
  isbn = {978-3-031-20082-3}
}

@article{fuMMEComprehensiveEvaluation2024,
  title = {MME: A Comprehensive Evaluation Benchmark for Multimodal Large Language Models},
  shorttitle = {MME},
  author = {Fu, Chaoyou and Chen, Peixian and Shen, Yunhang and Qin, Yulei and Zhang, Mengdan and Lin, Xu and Yang, Jinrui and Zheng, Xiawu and Li, Ke and Sun, Xing and Wu, Yunsheng and Ji, Rongrong},
  year = 2024,
  primaryclass = {cs},
  doi = {10.48550/arXiv.2306.13394},
  archiveprefix = {arXiv},
  journal = {arXiv preprint arXiv:2306.13394}
}

@article{geminiteamGemini15Unlocking2024,
  title = {Gemini 1.5: Unlocking Multimodal Understanding across Millions of Tokens of Context},
  shorttitle = {Gemini 1.5},
  author = {Gemini Team},
  year = 2024,
  doi = {10.48550/arXiv.2403.05530},
  archiveprefix = {arXiv},
  journal = {arXiv preprint arXiv:2403.05530}
}

@article{geminiteamGeminiFamilyHighly2024,
  title = {Gemini: A Family of Highly Capable Multimodal Models},
  shorttitle = {Gemini},
  author = {Gemini Team},
  year = 2024,
  doi = {10.48550/arXiv.2312.11805},
  archiveprefix = {arXiv},
  journal = {arXiv preprint arXiv:2312.11805}
}

@inproceedings{goldenGenerativeAILLMs2024,
  title = {Generative AI Beyond LLMs: System Implications of Multi-Modal Generation},
  shorttitle = {Generative AI Beyond LLMs},
  booktitle = {ISPASS},
  author = {Golden, Alicia and Hsia, Samuel and Sun, Fei and Acun, Bilge and Hosmer, Basil and Lee, Yejin and DeVito, Zachary and Johnson, Jeff and Wei, Gu-Yeon and Brooks, David and Wu, Carole-Jean},
  year = 2024,
  pages = {257--267},
  issn = {2766-0486},
  doi = {10.1109/ISPASS61541.2024.00032}
}

@article{gongMultiModalGPTVisionLanguage2023,
  title = {MultiModal-GPT: A Vision and Language Model for Dialogue with Humans},
  shorttitle = {MultiModal-GPT},
  author = {Gong, Tao and Lyu, Chengqi and Zhang, Shilong and Wang, Yudong and Zheng, Miao and Zhao, Qian and Liu, Kuikun and Zhang, Wenwei and Luo, Ping and Chen, Kai},
  year = 2023,
  primaryclass = {cs},
  archiveprefix = {arXiv},
  journal = {arXiv preprint arXiv:2305.04790}
}

@inproceedings{gurariVizWizGrandChallenge2018,
  title = {VizWiz Grand Challenge: Answering Visual Questions From Blind People},
  shorttitle = {VizWiz},
  booktitle = {CVPR},
  author = {Gurari, Danna and Li, Qing and Stangl, Abigale J. and Guo, Anhong and Lin, Chi and Grauman, Kristen and Luo, Jiebo and Bigham, Jeffrey P.},
  year = 2018,
  pages = {3608--3617}
}

@inproceedings{huangIVTPInstructionguidedVisual2024,
  title = {IVTP: Instruction-Guided Visual Token Pruning for Large Vision-Language Models},
  shorttitle = {IVTP},
  booktitle = {ECCV},
  author = {Huang, Kai and Zou, Hao and Xi, Ye and Wang, BoChen and Xie, Zhen and Yu, Liang},
  year = 2024,
  pages = {214--230}
}

@inproceedings{huangLanguageNotAll2023,
  title = {Language Is Not All You Need: Aligning Perception with Language Models},
  shorttitle = {Kosmos1},
  booktitle = {NeurIPS},
  author = {Huang, Shaohan and Dong, Li and Wang, Wenhui and Hao, Yaru and Singhal, Saksham and Ma, Shuming and Lv, Tengchao and Cui, Lei and Mohammed, Owais Khan and Patra, Barun and Liu, Qiang and Aggarwal, Kriti and Chi, Zewen and Bjorck, Nils and Chaudhary, Vishrav and Som, Subhojit and Song, Xia and Wei, Furu},
  year = 2023,
  pages = {72096--72109}
}

@inproceedings{hudsonGQANewDataset2019,
  title = {GQA: A New Dataset for Real-World Visual Reasoning and Compositional Question Answering},
  shorttitle = {GQA},
  booktitle = {CVPR},
  author = {Hudson, Drew A. and Manning, Christopher D.},
  year = 2019,
  pages = {6693--6702},
  issn = {2575-7075},
  doi = {10.1109/CVPR.2019.00686}
}

@inproceedings{huMatryoshkaQueryTransformer2025,
  title = {Matryoshka Query Transformer for Large Vision-Language Models},
  shorttitle = {MQT},
  booktitle = {NeurIPS},
  author = {Hu, Wenbo and Dou, Zi-Yi and Li, Liunian and Kamath, Amita and Peng, Nanyun and Chang, Kai-Wei},
  year = 2025,
  pages = {50168--50188}
}

@article{jiangMistral7B2023,
  title = {Mistral 7B},
  author = {Jiang, Albert Q. and Sablayrolles, Alexandre and Mensch, Arthur and Bamford, Chris and Chaplot, Devendra Singh and de las Casas, Diego and Bressand, Florian and Lengyel, Gianna and Lample, Guillaume and Saulnier, Lucile and Lavaud, L{\'e}lio Renard and Lachaux, Marie-Anne and Stock, Pierre and Scao, Teven Le and Lavril, Thibaut and Wang, Thomas and Lacroix, Timoth{\'e}e and Sayed, William El},
  year = 2023,
  primaryclass = {cs},
  doi = {10.48550/arXiv.2310.06825},
  archiveprefix = {arXiv},
  journal = {arXiv preprint arXiv:2310.06825}
}

@inproceedings{kimTokenFusionBridging2024,
  title = {Token Fusion: Bridging the Gap between Token Pruning and Token Merging},
  shorttitle = {Token Fusion},
  booktitle = {WACV},
  author = {Kim, Minchul and Gao, Shangqian and Hsu, Yen-Chang and Shen, Yilin and Jin, Hongxia},
  year = 2024,
  pages = {1372--1381},
  address = {Waikoloa, HI, USA},
  doi = {10.1109/WACV57701.2024.00141},
  copyright = {https://doi.org/10.15223/policy-029},
  isbn = {979-8-3503-1892-0}
}

@inproceedings{kwonEfficientMemoryManagement2023,
  title = {Efficient Memory Management for Large Language Model Serving with PagedAttention},
  booktitle = {SOSP},
  author = {Kwon, Woosuk and Li, Zhuohan and Zhuang, Siyuan and Sheng, Ying and Zheng, Lianmin and Yu, Cody Hao and Gonzalez, Joseph and Zhang, Hao and Stoica, Ion},
  year = 2023,
  pages = {611--626},
  doi = {10.1145/3600006.3613165},
  isbn = {979-8-4007-0229-7}
}

@inproceedings{leeInfiniGenEfficientGenerative2024,
  title = {InfiniGen: Efficient Generative Inference of Large Language Models with Dynamic KV Cache Management},
  shorttitle = {InfiniGen},
  booktitle = {OSDI},
  author = {Lee, Wonbeom and Lee, Jungi and Seo, Junghwan and Sim, Jaewoong},
  year = 2024,
  pages = {155--172},
  isbn = {978-1-939133-40-3}
}

@inproceedings{liBLIP2BootstrappingLanguageImage2023,
  title = {BLIP-2: Bootstrapping Language-Image Pre-Training with Frozen Image Encoders and Large Language Models},
  shorttitle = {BLIP-2},
  booktitle = {ICML},
  author = {Li, Junnan and Li, Dongxu and Savarese, Silvio and Hoi, Steven},
  year = 2023,
  pages = {19730--19742},
  issn = {2640-3498}
}

@inproceedings{liEvaluatingObjectHallucination2023,
  title = {Evaluating Object Hallucination in Large Vision-Language Models},
  shorttitle = {POPE},
  booktitle = {EMNLP},
  author = {Li, Yifan and Du, Yifan and Zhou, Kun and Wang, Jinpeng and Zhao, Xin and Wen, Ji-Rong},
  editor = {Bouamor, Houda and Pino, Juan and Bali, Kalika},
  year = 2023,
  pages = {292--305},
  doi = {10.18653/v1/2023.emnlp-main.20}
}

@inproceedings{liLLaMAVIDImageWorth2024,
  title = {LLaMA-VID: An Image Is Worth 2 Tokens in~Large Language Models},
  shorttitle = {LLaMA-VID},
  booktitle = {ECCV},
  author = {Li, Yanwei and Wang, Chengyao and Jia, Jiaya},
  editor = {Leonardis, Ale{\v s} and Ricci, Elisa and Roth, Stefan and Russakovsky, Olga and Sattler, Torsten and Varol, G{\"u}l},
  year = 2024,
  pages = {323--340},
  doi = {10.1007/978-3-031-72952-2_19},
  isbn = {978-3-031-72952-2}
}

@article{liLLaVAOneVisionEasyVisual2024,
  title = {LLaVA-OneVision: Easy Visual Task Transfer},
  shorttitle = {LLaVA-OneVision},
  author = {Li, Bo and Zhang, Yuanhan and Guo, Dong and Zhang, Renrui and Li, Feng and Zhang, Hao and Zhang, Kaichen and Zhang, Peiyuan and Li, Yanwei and Liu, Ziwei and Li, Chunyuan},
  year = 2024,
  primaryclass = {cs},
  doi = {10.48550/arXiv.2408.03326},
  archiveprefix = {arXiv},
  journal = {arXiv preprint arXiv:2408.03326}
}

@inproceedings{linMicrosoftCOCOCommon2014,
  title = {Microsoft COCO: Common Objects in Context},
  shorttitle = {Microsoft COCO},
  booktitle = {ECCV},
  author = {Lin, Tsung-Yi and Maire, Michael and Belongie, Serge and Hays, James and Perona, Pietro and Ramanan, Deva and Doll{\'a}r, Piotr and Zitnick, C. Lawrence},
  editor = {Fleet, David and Pajdla, Tomas and Schiele, Bernt and Tuytelaars, Tinne},
  year = 2014,
  pages = {740--755},
  address = {Cham},
  doi = {10.1007/978-3-319-10602-1_48},
  isbn = {978-3-319-10602-1}
}

@article{liOtterHDHighResolutionMultimodality2023,
  title = {OtterHD: A High-Resolution Multi-Modality Model},
  shorttitle = {OtterHD},
  author = {Li, Bo and Zhang, Peiyuan and Yang, Jingkang and Zhang, Yuanhan and Pu, Fanyi and Liu, Ziwei},
  year = 2023,
  primaryclass = {cs},
  doi = {10.48550/arXiv.2311.04219},
  archiveprefix = {arXiv},
  journal = {arXiv preprint arXiv:2311.04219}
}

@inproceedings{liSEEDBenchBenchmarkingMultimodal2024,
  title = {SEED-Bench: Benchmarking Multimodal Large Language Models},
  shorttitle = {SEED-Bench},
  booktitle = {CVPR},
  author = {Li, Bohao and Ge, Yuying and Ge, Yixiao and Wang, Guangzhi and Wang, Rui and Zhang, Ruimao and Shan, Ying},
  year = 2024,
  pages = {13299--13308},
  issn = {2575-7075},
  doi = {10.1109/CVPR52733.2024.01263}
}

@article{liSemanticRoutingExploring2026,
  title = {Semantic Routing: Exploring Multi-Layer LLM Feature Weighting for Diffusion Transformers},
  shorttitle = {Semantic Routing},
  author = {Li, Bozhou and Guan, Yushuo and Li, Haolin and Zeng, Bohan and Ji, Yiyan and Ding, Yue and Wan, Pengfei and Gai, Kun and Zhang, Yuanxing and Zhang, Wentao},
  year = 2026,
  primaryclass = {cs},
  doi = {10.48550/arXiv.2602.03510},
  archiveprefix = {arXiv},
  journal = {arXiv preprint arXiv:2602.03510}
}

@inproceedings{liSnapKVLLMKnows2024,
  title = {SnapKV: LLM Knows What You Are Looking for Before Generation},
  shorttitle = {SnapKV},
  booktitle = {NeurIPS},
  author = {Li, Yuhong and Huang, Yingbing and Yang, Bowen and Venkitesh, Bharat and Locatelli, Acyr and Ye, Hanchen and Cai, Tianle and Lewis, Patrick and Chen, Deming},
  year = 2024,
  pages = {22947--22970}
}

@inproceedings{liuEfficientInferenceVision2024,
  title = {Efficient Inference of~Vision Instruction-Following Models with~Elastic Cache},
  booktitle = {ECCV},
  author = {Liu, Zuyan and Liu, Benlin and Wang, Jiahui and Dong, Yuhao and Chen, Guangyi and Rao, Yongming and Krishna, Ranjay and Lu, Jiwen},
  editor = {Leonardis, Ale{\v s} and Ricci, Elisa and Roth, Stefan and Russakovsky, Olga and Sattler, Torsten and Varol, G{\"u}l},
  year = 2024,
  pages = {54--69},
  doi = {10.1007/978-3-031-72643-9_4},
  isbn = {978-3-031-72643-9}
}

@article{liuImprovedBaselinesVisual2024,
  title = {Improved Baselines with Visual Instruction Tuning},
  shorttitle = {LLaVA-1.5},
  author = {Liu, Haotian and Li, Chunyuan and Li, Yuheng and Lee, Yong Jae},
  year = 2024,
  primaryclass = {cs},
  doi = {10.48550/arXiv.2310.03744},
  archiveprefix = {arXiv},
  journal = {arXiv preprint arXiv:2310.03744}
}

@misc{liuLLaVANeXTImprovedReasoning2024,
  title = {LLaVA-NeXT: Improved Reasoning, OCR, and World Knowledge},
  shorttitle = {LLaVA-NeXT},
  author = {Liu, Haotian and Li, Chunyuan and Li, Yuheng and Li, Bo and Zhang, Yuanhan and Shen, Sheng and Lee, Yong Jae},
  year = 2024,
  howpublished = {https://llava-vl.github.io/blog/2024-01-30-llava-next/}
}

@inproceedings{liuMiniCacheKVCache2024,
  title = {MiniCache: KV Cache Compression in Depth Dimension for Large Language Models},
  shorttitle = {MiniCache},
  booktitle = {NeurIPS},
  author = {Liu, Akide and Liu, Jing and Pan, Zizheng and He, Yefei and Haffari, Gholamreza and Zhuang, Bohan},
  year = 2024,
  pages = {139997--140031}
}

@inproceedings{liuScissorhandsExploitingPersistence2023,
  title = {Scissorhands: Exploiting the Persistence of Importance Hypothesis for LLM KV Cache Compression at Test Time},
  shorttitle = {Scissorhands},
  booktitle = {NeurIPS},
  author = {Liu, Zichang and Desai, Aditya and Liao, Fangshuo and Wang, Weitao and Xie, Victor and Xu, Zhaozhuo and Kyrillidis, Anastasios and Shrivastava, Anshumali},
  year = 2023,
  pages = {52342--52364}
}

@inproceedings{liuSwinTransformerHierarchical2021,
  title = {Swin Transformer: Hierarchical Vision Transformer Using Shifted Windows},
  shorttitle = {Swin Transformer},
  booktitle = {ICCV},
  author = {Liu, Ze and Lin, Yutong and Cao, Yue and Hu, Han and Wei, Yixuan and Zhang, Zheng and Lin, Stephen and Guo, Baining},
  year = 2021,
  pages = {10012--10022}
}

@inproceedings{liuVisualInstructionTuning2023,
  title = {Visual Instruction Tuning},
  shorttitle = {Llava},
  booktitle = {NeurIPS},
  author = {Liu, Haotian and Li, Chunyuan and Wu, Qingyang and Lee, Yong Jae},
  year = 2023,
  pages = {34892--34916}
}

@article{llamateamLlama3Herd2024,
  title = {The Llama 3 Herd of Models},
  shorttitle = {Llama3},
  author = {{Llama Team}},
  year = 2024,
  doi = {10.48550/arXiv.2407.21783},
  archiveprefix = {arXiv},
  journal = {arXiv preprint arXiv:2407.21783v3}
}

@inproceedings{luContentawareTokenSharing2023,
  title = {Content-Aware Token Sharing for Efficient Semantic Segmentation with Vision Transformers},
  booktitle = {CVPR},
  author = {Lu, Chenyang and {de Geus}, Daan and Dubbelman, Gijs},
  year = 2023,
  pages = {23631--23640},
  issn = {2575-7075},
  doi = {10.1109/CVPR52729.2023.02263}
}

@article{luDeepSeekVLRealWorldVisionLanguage2024,
  title = {DeepSeek-VL: Towards Real-World Vision-Language Understanding},
  shorttitle = {DeepSeek-VL},
  author = {Lu, Haoyu and Liu, Wen and Zhang, Bo and Wang, Bingxuan and Dong, Kai and Liu, Bo and Sun, Jingxiang and Ren, Tongzheng and Li, Zhuoshu and Yang, Hao and Sun, Yaofeng and Deng, Chengqi and Xu, Hanwei and Xie, Zhenda and Ruan, Chong},
  year = 2024,
  primaryclass = {cs},
  doi = {10.48550/arXiv.2403.05525},
  archiveprefix = {arXiv},
  journal = {arXiv preprint arXiv:2403.05525}
}

@inproceedings{luLearnExplainMultimodal2022,
  title = {Learn to Explain: Multimodal Reasoning via Thought Chains for Science Question Answering},
  shorttitle = {ScienceQA},
  booktitle = {NeurIPS},
  author = {Lu, Pan and Mishra, Swaroop and Xia, Tony and Qiu, Liang and Chang, Kai-Wei and Zhu, Song-Chun and Tafjord, Oyvind and Clark, Peter and Kalyan, Ashwin},
  year = 2022,
  pages = {2507--2521}
}

@article{maoPruneMergeEfficient2025,
  title = {Prune and Merge: Efficient Token Compression for Vision Transformer with Spatial Information Preserved},
  shorttitle = {Prune and Merge},
  author = {Mao, Junzhu and Shen, Yang and Guo, Jinyang and Yao, Yazhou and Hua, Xiansheng and Shen, Hengtao},
  year = 2025,
  journal = {TMM},
  pages = {1--14},
  issn = {1941-0077},
  doi = {10.1109/TMM.2025.3535405}
}

@inproceedings{norouziALGMAdaptiveLocalthenGlobal2024,
  title = {ALGM: Adaptive Local-Then-Global Token Merging for Efficient Semantic Segmentation with Plain Vision Transformers},
  shorttitle = {ALGM},
  booktitle = {CVPR},
  author = {Norouzi, Narges and Orlova, Svetlana and De Geus, Daan and Dubbelman, Gijs},
  year = 2024,
  pages = {15773--15782},
  issn = {2575-7075},
  doi = {10.1109/CVPR52733.2024.01493}
}

@article{openaiGPT4TechnicalReport2024,
  title = {GPT-4 Technical Report},
  shorttitle = {GPT-4},
  author = {OpenAI},
  year = 2024,
  doi = {10.48550/arXiv.2303.08774},
  archiveprefix = {arXiv},
  journal = {arXiv preprint arXiv:2303.08774}
}

@inproceedings{panLessMorePay2022,
  title = {Less Is More: Pay Less Attention in Vision Transformers},
  shorttitle = {Less Is More},
  booktitle = {AAAI},
  author = {Pan, Zizheng and Zhuang, Bohan and He, Haoyu and Liu, Jing and Cai, Jianfei},
  year = 2022,
  volume = {36},
  pages = {2035--2043},
  doi = {10.1609/aaai.v36i2.20099}
}

@inproceedings{popeEfficientlyScalingTransformer2023,
  title = {Efficiently Scaling Transformer Inference},
  booktitle = {MLSys},
  author = {Pope, Reiner and Douglas, Sholto and Chowdhery, Aakanksha and Devlin, Jacob and Bradbury, James and Levskaya, Anselm and Heek, Jonathan and Xiao, Kefan and Agrawal, Shivani and Dean, Jeff},
  year = 2023,
  pages = {606--624}
}

@inproceedings{radfordLearningTransferableVisual2021,
  title = {Learning Transferable Visual Models From Natural Language Supervision},
  booktitle = {ICML},
  author = {Radford, Alec and Kim, Jong Wook and Hallacy, Chris and Ramesh, Aditya and Goh, Gabriel and Agarwal, Sandhini and Sastry, Girish and Askell, Amanda and Mishkin, Pamela and Clark, Jack and Krueger, Gretchen and Sutskever, Ilya},
  year = 2021,
  pages = {8748--8763},
  issn = {2640-3498}
}

@inproceedings{raoDynamicViTEfficientVision2021,
  title = {DynamicViT: Efficient Vision Transformers with Dynamic Token Sparsification},
  shorttitle = {DynamicViT},
  booktitle = {NeurIPS},
  author = {Rao, Yongming and Zhao, Wenliang and Liu, Benlin and Lu, Jiwen and Zhou, Jie and Hsieh, Cho-Jui},
  year = 2021,
  pages = {13937--13949}
}

@inproceedings{songHierarchicalContextMerging2024,
  title = {Hierarchical Context Merging: Better Long Context Understanding for Pre-Trained LLMs},
  shorttitle = {Hierarchical Context Merging},
  booktitle = {ICLR},
  author = {Song, Woomin and Oh, Seunghyuk and Mo, Sangwoo and Kim, Jaehyung and Yun, Sukmin and Ha, Jung-Woo and Shin, Jinwoo},
  year = 2024
}

@misc{thevicunateamVicunaOpenSourceChatbot2023,
  title = {Vicuna: An Open-Source Chatbot Impressing GPT-4 with 90\% ChatGPT Quality},
  shorttitle = {Vicuna},
  author = {{Vicuna Team}},
  year = 2023,
  howpublished = {https://lmsys.org/blog/2023-03-30-vicuna}
}

@inproceedings{tranAcceleratingTransformersSpectrumPreserving2024,
  title = {Accelerating Transformers with Spectrum-Preserving Token Merging},
  shorttitle = {PiToMe},
  booktitle = {NeurIPS},
  author = {Tran, Hoai-Chau and Nguyen, Duy M. and Nguyen, TrungTin and Le, Ngan and Xie, Pengtao and Sonntag, Daniel and Zou, James and Nguyen, Binh T. and Niepert, Mathias},
  year = 2024,
  pages = {30772--30810}
}

@inproceedings{vaswaniAttentionAllYou2017,
  title = {Attention Is All You Need},
  booktitle = {NeurIPS},
  author = {Vaswani, Ashish and Shazeer, Noam and Parmar, Niki and Uszkoreit, Jakob and Jones, Llion and Gomez, Aidan N and ukasz Kaiser, {\L} and Polosukhin, Illia},
  year = 2017,
  volume = {30},
  pages = {6000--6010}
}

@article{wangEnhancingReasoningAbility2025,
  title = {Enhancing the Reasoning Ability of Multimodal Large Language Models via Mixed Preference Optimization},
  author = {Wang, Weiyun and Chen, Zhe and Wang, Wenhai and Cao, Yue and Liu, Yangzhou and Gao, Zhangwei and Zhu, Jinguo and Zhu, Xizhou and Lu, Lewei and Qiao, Yu and Dai, Jifeng},
  year = 2025,
  primaryclass = {cs},
  doi = {10.48550/arXiv.2411.10442},
  archiveprefix = {arXiv},
  journal = {arXiv preprint arXiv:2411.10442}
}

@article{wangUnderstandingMitigatingMiscalibration2024,
  title = {Understanding and Mitigating Miscalibration in Prompt Tuning for Vision-Language Models},
  author = {Wang, Shuoyuan and Li, Yixuan and Wei, Hongxin},
  year = 2024,
  primaryclass = {cs},
  doi = {10.48550/arXiv.2410.02681},
  archiveprefix = {arXiv},
  journal = {arXiv preprint arXiv:2410.02681}
}

@inproceedings{weiJointTokenPruning2023,
  title = {Joint Token Pruning and Squeezing Towards More Aggressive Compression of Vision Transformers},
  booktitle = {CVPR},
  author = {Wei, Siyuan and Ye, Tianzhu and Zhang, Shen and Tang, Yao and Liang, Jiajun},
  year = 2023,
  pages = {2092--2101},
  issn = {2575-7075},
  doi = {10.1109/CVPR52729.2023.00208}
}

@inproceedings{xuFastOndeviceLLM2025,
  title = {Fast On-Device LLM Inference with NPUs},
  booktitle = {ASPLOS},
  author = {Xu, Daliang and Zhang, Hao and Yang, Liming and Liu, Ruiqi and Huang, Gang and Xu, Mengwei and Liu, Xuanzhe},
  year = 2025,
  series = {ASPLOS '25},
  pages = {445--462},
  doi = {10.1145/3669940.3707239},
  isbn = {979-8-4007-0698-1}
}

@inproceedings{xuGroupViTSemanticSegmentation2022,
  title = {GroupViT: Semantic Segmentation Emerges from Text Supervision},
  shorttitle = {GroupViT},
  booktitle = {CVPR},
  author = {Xu, Jiarui and De Mello, Shalini and Liu, Sifei and Byeon, Wonmin and Breuel, Thomas and Kautz, Jan and Wang, Xiaolong},
  year = 2022,
  pages = {18113--18123},
  doi = {10.1109/CVPR52688.2022.01760},
  copyright = {https://doi.org/10.15223/policy-029},
  isbn = {978-1-6654-6946-3}
}

@inproceedings{yangPyramidInferPyramidKV2024,
  title = {PyramidInfer: Pyramid KV Cache Compression for High-Throughput LLM Inference},
  shorttitle = {PyramidInfer},
  booktitle = {ACL Findings},
  author = {Yang, Dongjie and Han, Xiaodong and Gao, Yan and Hu, Yao and Zhang, Shilin and Zhao, Hai},
  editor = {Ku, Lun-Wei and Martins, Andre and Srikumar, Vivek},
  year = 2024,
  pages = {3258--3270},
  doi = {10.18653/v1/2024.findings-acl.195}
}

@inproceedings{yinAViTAdaptiveTokens2022,
  title = {A-ViT: Adaptive Tokens for Efficient Vision Transformer},
  shorttitle = {A-ViT},
  booktitle = {CVPR},
  author = {Yin, Hongxu and Vahdat, Arash and Alvarez, Jose M. and Mallya, Arun and Kautz, Jan and Molchanov, Pavlo},
  year = 2022,
  pages = {10809--10818}
}

@inproceedings{yuMMVetEvaluatingLarge2024,
  title = {MM-Vet: Evaluating Large Multimodal Models for Integrated Capabilities},
  shorttitle = {MM-Vet},
  booktitle = {ICML},
  author = {Yu, Weihao and Yang, Zhengyuan and Li, Linjie and Wang, Jianfeng and Lin, Kevin and Liu, Zicheng and Wang, Xinchao and Wang, Lijuan},
  year = 2024,
  volume = {235},
  pages = {57730--57754}
}

@inproceedings{zhangH2OHeavyHitterOracle2023,
  title = {H2O: Heavy-Hitter Oracle for Efficient Generative Inference of Large Language Models},
  shorttitle = {H2O},
  booktitle = {NeurIPS},
  author = {Zhang, Zhenyu and Sheng, Ying and Zhou, Tianyi and {Tianlong Chen} and Zheng, Lianmin and Cai, Ruisi and Song, Zhao and Tian, Yuandong and Re, Christopher and Barrett, Clark and Wang, Zhangyang and Chen, Beidi},
  year = 2023,
  pages = {34661--34710}
}

@inproceedings{zhangLLaVAMiniEfficientImage2024,
  title = {LLaVA-Mini: Efficient Image and Video Large Multimodal Models with One Vision Token},
  shorttitle = {LLaVA-Mini},
  booktitle = {ICLR},
  author = {Zhang, Shaolei and Fang, Qingkai and Yang, Zhe and Feng, Yang},
  year = 2024
}

@article{zhangLMMsEvalRealityCheck2025,
  title = {LMMs-Eval: Reality Check on the Evaluation of Large Multimodal Models},
  shorttitle = {LMMs-Eval},
  author = {Zhang, Kaichen and Li, Bo and Zhang, Peiyuan and Pu, Fanyi and Cahyono, Joshua Adrian and Hu, Kairui and Liu, Shuai and Zhang, Yuanhan and Yang, Jingkang and Li, Chunyuan and Liu, Ziwei},
  year = 2025,
  primaryclass = {cs},
  doi = {10.48550/arXiv.2407.12772},
  archiveprefix = {arXiv},
  journal = {arXiv preprint arXiv:2407.12772}
}

@inproceedings{zhangVideoLLaMAInstructiontunedAudioVisual2023,
  title = {Video-LLaMA: An Instruction-Tuned Audio-Visual Language Model for Video Understanding},
  shorttitle = {Video-LLaMA},
  booktitle = {EMNLP},
  author = {Zhang, Hang and Li, Xin and Bing, Lidong},
  editor = {Feng, Yansong and Lefever, Els},
  year = 2023,
  pages = {543--553},
  doi = {10.18653/v1/2023.emnlp-demo.49}
}

@inproceedings{zhuMiniGPT4EnhancingVisionLanguage2024,
  title = {MiniGPT-4: Enhancing Vision-Language Understanding with Advanced Large Language Models},
  shorttitle = {MiniGPT-4},
  booktitle = {ICLR},
  author = {Zhu, Deyao and Chen, Jun and Shen, Xiaoqian and Li, Xiang and Elhoseiny, Mohamed},
  year = 2024
}
